\documentclass[10pt]{article} % For LaTeX2e
\usepackage[accepted]{rlc}

\usepackage{amssymb}            % Defines common symbols like \mathbb R
\usepackage{mathtools}          % Extends amsmath, providing common math tools
\usepackage{mathrsfs}           % Enables \mathscr, which can work in cases that \mathcal does not
\usepackage{graphicx}           % For including images
\usepackage{subcaption}         % Allows for the use of subfigures and subcaptions
\usepackage[space]{grffile}     % For spaces in image names
\usepackage{url}                % For displaying urls
\usepackage{wrapfig}
\usepackage{booktabs}
\usepackage{rotating}
\usepackage{cuted}

\usepackage{amsmath}
\usepackage{amsfonts}
\usepackage{amsthm}
\usepackage{amssymb}
\usepackage{mathtools}
\usepackage{tcolorbox}
\usepackage{algorithm}
\usepackage{algpseudocode}

\theoremstyle{plain}
\newtheorem{theorem}{Theorem}%[section]

\newtheorem{corollary}[theorem]{Corollary}

\theoremstyle{definition}

\theoremstyle{remark}

\newtheorem*{theorem*}{Theorem}
\newtheorem*{lemma*}{Lemma}
\newtheorem*{definition*}{Definition}
\newtheorem*{corollary*}{Corollary}
\newtheorem*{remark*}{Remark}

\DeclareMathOperator*{\E}{\mathbb{E}}

\newcommand{\s}{\mathcal{S}}

\newcommand{\A}{\mathcal{A}}

\newcommand{\T}{\mathcal{T}}

\usepackage{color}
\usepackage{multicol}
\usepackage[normalem]{ulem} % Include this in the preamble to use \uline

\title{Boosting Soft Q-Learning by Bounding}
\author{Jacob Adamczyk  \\
    jacob.adamczyk001@umb.edu \\
    Department of Physics \\
    University of Massachusetts Boston
    \And 
    Volodymyr Makarenko  \\
    volodymyr.makarenko@sjsu.edu \\
    Department of Computer Engineering \\
    San Jos\'e State University
    \And
    Stas Tiomkin   \\
    stas.tiomkin@sjsu.edu \\
    Department of Computer Engineering \\
    San Jos\'e State University
    \And
    Rahul V. Kulkarni   \\
    rahul.kulkarni@umb.edu\\
    Department of Physics \\
    University of Massachusetts Boston
    }

\begin{document}

%%%%%%%%%%%%%%%
%%%% VOCAB %%%%
%%%%%%%%%%%%%%%
\def\CaoThm{Theorem~1 }
\def\CaoThmNospace{Theorem~1}
\def\HaarLem{Lemma 1 }
\def\zeroshot{zero-shot}
%%%%%%%%%%%%%%%%
%%% THEOREMS %%%
%%%%%%%%%%%%%%%%

\newcommand\av[1]{\left\lvert#1\right\rvert}
\newcommand{\fr}{f\left(\{r_j (s,a)\}\right)}
\newcommand{\fQ}{f\left(\{Q_j^* (s,a)\}\right)}
\newcommand{\G}{\mathcal{G}}

\newcommand{\fQspap}{f\left(\{Q_j^* (s',a')\}\right)}
%% For appendix custom numbering:
\theoremstyle{plain}
\newtheorem{innercustomgeneric}{\customgenericname}
\providecommand{\customgenericname}{}
\newcommand{\newcustomtheorem}[2]{%
  \newenvironment{#1}[1]
  {%
   \renewcommand\customgenericname{#2}%
   \renewcommand\theinnercustomgeneric{##1}%
   \innercustomgeneric
  }
  {\endinnercustomgeneric}
}

\newcustomtheorem{customthm}{Theorem}
\newcustomtheorem{customprop}{Proposition}
\newcustomtheorem{customlemma}{Lemma}
\newcustomtheorem{customcor}{Corollary}

\def\AAAITheorem#1{
    \begin{customthm}{5.1}[\cite{Adamczyk_AAAI}]
        Let a set of primitive tasks $\{\T_j\}_{j=1}^{M}$
        with corresponding optimal value functions $\{Q_j^*\}$ be given.
        Denote $\widetilde{Q}^*$ as the optimal action-value function for the task $\widetilde{\T}$ with reward function $\widetilde{r}(s,a)$. Define $K^*$ as the optimal soft action-value function for a task with reward function $\kappa$ and prior policy $\pi_f$ with the following definitions:
        \begin{align*}\label{eq:rwd for composition}
            \kappa(s,a) = \widetilde{r}(s,a) + \gamma \E_{s' \sim{} p} V_f(s') - \fQ,
        \end{align*}
        \begin{equation*}
            V_f(s) = \frac{1}{\beta}\log\E_{a\sim{} \pi_0}\exp{\beta \fQ },
        \end{equation*}
        % and $\pi_f$ is the policy derived from $\fQ$:
        \begin{equation}
            \pi_f(a\vert s) = \pi_0(a\vert s) \frac{e^{\beta \fQ }}{e^{\beta V_f(s)}}.
        \end{equation}
        Then, the optimal value function $\widetilde{Q}^*$ and the \zeroshot\ solution $f(\{Q_j^*\})$ are related by:
        \begin{equation}
            \widetilde{Q}^*(s,a) = \fQ + K^*(s,a)
            \label{eq:qtilda=f(q)+K}
        \end{equation}
        \label{thm:aaai#1}
    \end{customthm}
}

\def\StdRLRwdChange#1{
    \begin{customthm}{6.1a}[\citeauthor{ng_shaping}]
        Let a (standard RL) primitive task $\T$ with reward function $r$ be given, with the optimal value function $V^*(s)$.
        Consider another (standard RL) task, $\widetilde{\T}$ with reward function $\widetilde{r}$, with an unknown optimal action-value function, $\widetilde{Q}^*$.
        Define $\kappa(s,a) \doteq \widetilde{r}(s,a) + \gamma \E_{s'} V^*(s') - V^*(s)$.\\
        Denote the optimal action-value function $K^*$ as the solution of the following Bellman optimality equation
        \begin{equation}\label{eq:K_backup1#1}
            K^{*}(s,a) = \kappa(s,a) + \gamma \E_{s' \sim{} p} \max_{a'} K^*(s',a')
        \end{equation}
        Then, \begin{equation}\label{eq:Q=Q+K rwd_change#1}
            \widetilde{Q}^*(s,a) = V^*(s) + K^*(s,a)
        \end{equation}
        \label{thm:rwd_change_stdRL#1}
    \end{customthm}
}

\def\StdRLComp#1{
    \begin{customthm}{4.8}
        Given a set of primitive tasks $\{\T_j\}$
        with corresponding optimal value functions $\{Q_j^*\}$, denote
        $\widetilde{Q}^*$ as the optimal value function for the composition
        of $\{\T_j\}$ under the composition function $f: \mathbb{R}^M \to \mathbb{R}$.

        Define $K^*$ as the optimal value function for a task with reward function $\kappa$ defined by:
        \begin{align*}\label{eq:rwd for composition}
            \kappa(s,a) = f(\{r_j(s,a)\}) + \gamma \E_{s'} V_f(s') - V_f(s)
        \end{align*}
        \begin{equation*}
            V_f(s) = \max_a \fQ
        \end{equation*}

        Then, the optimal value functions $\widetilde{Q}^*$ and $K^*$ are related by:
        \begin{equation}
            \widetilde{Q}^*(s,a) = V_f(s) + K^*(s,a)
            \label{eq:std RL qtilda=f(q)+K#1}
        \end{equation}
        \label{thm:std_aaai_#1}
    \end{customthm}
}

\def\ClippingCorollaryExact#1{
    \begin{customcor}{5.2}[Zero-Shot Bounds]
        Let a bounded composition function $f: \mathbb{R}^M \to \mathbb{R}$ be given. The optimal value function
        for the composite task, $\widetilde{Q}^*(s,a)$, is upper- and lower-bounded at the state-action level:
        % \begin{equation}
        %     \frac{\gamma \min\kappa}{1-\gamma} \leq \widetilde{Q}^*(s,a) - \fQ \leq \frac{\max \kappa}{1-\gamma}.
        % \end{equation}\label{eq:exact_bound}
        % \begin{align}
        %     \widetilde{Q}^*(s,a) & \leq \fQ + \kappa(s,a) + \gamma \frac{\max \kappa}{1-\gamma} \\
        %     \widetilde{Q}^*(s,a) & \geq \fQ + \kappa(s,a) + \gamma \frac{\min\kappa}{1-\gamma}.
        % \end{align}\label{eq:exact_bound}
        \begin{align}
            \widetilde{Q}^*(s,a) & \leq \fr + \gamma \left(\E_{s'}V_f(s') + \frac{\max \kappa}{1-\gamma} \right) \label{eq:fq_K_bound1#1} \\
            \widetilde{Q}^*(s,a) & \geq \fr + \gamma \left(\E_{s'}V_f(s') + \frac{\min\kappa}{1-\gamma} \right) \label{eq:fq_K_bound2#1} 
        \end{align}
        \label{cor:clipping_exact#1}
    \end{customcor}
}

\def\SuboptimalityTheoremComp#1{
    \begin{customthm}{5.3}
        Let $\pi_f$ denote the \zeroshot\ policy obtained from the primitive value functions $\{Q^*_j\}$:
        \begin{equation}
            \pi_f(a\vert s) = \pi_0(a\vert s) \frac{e^{\beta \fQ }}{e^{\beta V_f(s)}}.
        \end{equation}

        Consider the soft value of the policy $\pi_f$ obtained from deploying it in the composite task:
        $\widetilde{Q}^{\pi_f}(s,a)$, and denote $\widetilde{V}^{\pi_f}=1/\beta \log \E_{a' \sim{} \pi_0} \exp \beta \fQ$.
        % Define the advantage functions $A_f = \fQ - V_f(s)$ and $\widetilde{A}^{\pi_f} = \widetilde{Q}^{\pi_f}(s,a) - \widetilde{V}^{\pi_f}(s)$.
        Let $\hat{D}^*$ be the optimal soft value function for the task with prior policy $\pi_f$ and reward function
        % \begin{equation}
        %     \hat{d}(s,a) = \gamma \E_{s' \sim{} p} \E_{a' \sim{} \pi_f} \left[ A_f(s',a') - \widetilde{A}^{\pi_f}(s',a') \right]
        % \end{equation}
        \begin{equation}
            \hat{d}(s,a) = \fr + \gamma \E_{s' \sim{} p} \widetilde{V}^{\pi_f}(s') - \widetilde{Q}^{\pi_f}(s,a)
        \end{equation}
        Then, the value functions satisfy:
        \begin{equation}
            \widetilde{Q}^{\pi_f}(s,a) = \widetilde{Q}^*(s,a) - \hat{D}^*(s,a)
        \end{equation}
        \label{thm:pi_f#1}\end{customthm}

}
\def\SuboptBound#1{
    \begin{customcor}{4.2}[Suboptimality Bounds]
        Let policy $\pi(a|s)$ be given with soft value $Q^{\pi}(s,a)$. The rate of the suboptimality gap, $Q^*(s,a) - Q^\pi(s,a)$, is then bounded between
        % \JA{replace qtilda with D:}
        \begin{equation}
        \inf_{(s,a)} d(s,a) \leq \frac{Q^*(s,a) - Q^\pi(s,a)}{H} \leq \sup_{(s,a)} d(s,a)
        \end{equation}

        where $d(s,a) \doteq r(s,a) + \gamma \E_{s'} V^{\pi}(s') - Q^{\pi}(s,a)$, $V^{\pi}(s) \doteq \log \E_{a} \exp \beta Q^{\pi}(s,a)$ is the soft state-value function, and $H = (1-\gamma)^{-1}$ is the effective time horizon.
        \label{cor:regret_bounds#1}
    \end{customcor}
}

\def\SuboptimalityTheoremHaarnoja#1{

    \begin{corollary}
        With the definitions of Theorem~\ref{thm:pi_f}, the value of $\pi_f$ satisfies
        \begin{equation}
            \widetilde{Q}^{\pi_f}(s,a) \geq \widetilde{Q}^*(s,a) - D^*(s,a)
        \end{equation}
        where $D^*(s,a)$ is the standard RL ($\beta \to \infty$ limit) optimal value function
        corresponding to the reward function $\hat{d}$ of Theorem \ref{thm:pi_f}.
    \end{corollary}

}

\def\OnTheFlyBoundComposite#1{
    \begin{customthm}{5.5}
        Let $\pi_k$ be the policy obtained at step $k$ during training of the composite value function, $\widetilde{Q}^*(s,a)$.

        Consider the corresponding soft-value of the policy $\pi_k$ obtained by deploying it in the composite task's environment:
        $\widetilde{Q}^{\pi_k}$, and denote $\widetilde{V}^{\pi_k}~=~1/\beta \log \E_{a \sim \pi_0} \exp \beta \widetilde{Q}^{\pi_k}(s,a)$.
        % Define the advantage functions $A_f = \fQ - V_f(s)$ and $\widetilde{A}^{\pi_k} = \widetilde{Q}^{\pi_k}(s,a) - \widetilde{V}^{\pi_k}(s)$
        Define the reward function
        % \begin{equation}
        %     \hat{d}_k(s,a) = \gamma \E_{s' \sim{} p} \E_{a' \sim{} \pi_f} \left[ A_f(s',a') - \widetilde{A}^{\pi_k}(s',a') \right].
        % \end{equation}
        \begin{equation}
            \hat{d}_k(s,a) = \fr + \gamma \E_{s' \sim{} p} \widetilde{V}^{\pi_k}(s') - \widetilde{Q}^{\pi_k}(s,a)
        \end{equation}
        Then the optimal value function for the composite task, $\widetilde{Q}^*$ is bounded by:
        % \begin{equation}
        %     \frac{\min{\hat{d}_k}}{1-\gamma} \leq \widetilde{Q}^*(s,a) - \widetilde{Q}^{\pi_k}(s,a) \leq \frac{\max{\hat{d}_k}}{1-\gamma}.
        %     \label{eq:pi_k#1}
        % \end{equation}
        \begin{align}
            \widetilde{Q}^*(s,a) & \leq \fr + \gamma \left(\E_{s' \sim{} p} \widetilde{V}^{\pi_k}(s') + \frac{\max{\hat{d}_k}}{1-\gamma}\right)  \\
            \widetilde{Q}^*(s,a) & \geq \fr + \gamma \left(\E_{s' \sim{} p} \widetilde{V}^{\pi_k}(s') + \frac{\min{\hat{d}_k}}{1-\gamma}\right)
            \label{eq:pi_k#1}
        \end{align}
        \label{thm:pi_k#1}
    \end{customthm}

}

\def\OnTheFlyBound#1{
    \begin{customthm}{1}[]
        Consider an entropy-regularized MDP $\langle \s, \A, p, r, \gamma, \beta, \pi_0 \rangle$ with optimal value function $Q^*(s,a)$.
        Let any bounded function $Q(s,a)$ be given.
        Denote the corresponding state-value function as $V(s)~\doteq~1/\beta \log \E_{a \sim \pi_0} \exp \beta Q(s,a)$. Then, $Q^*(s,a)$ is bounded by:
        % \begin{subequations}
        % \begin{align}
        %     Q^*(s,a) &\geq r(s,a) + \gamma \left(\E_{s' \sim{} p} V(s') + \frac{\inf \Delta}{1-\gamma}\right) \label{eq:gen_double_sided_boundsA#1}\\
        %     Q^*(s,a) &\leq r(s,a) + \gamma \left(\E_{s' \sim{} p} V(s') + \frac{\sup\Delta}{1-\gamma}\right) \label{eq:gen_double_sided_boundsB#1}
        % \end{align}
        % \end{subequations}
        \begin{equation}
        r(s,a) + \gamma \left(\E_{s' \sim{} p} V(s')+\frac{\inf \Delta}{1-\gamma}\right) \leq Q^*(s,a) \leq r(s,a) + \gamma \left(\E_{s' \sim{} p} V(s') + \frac{\sup\Delta}{1-\gamma}\right) \label{eq:gen_double_sided_bound#1}
        \end{equation}
        where         
        \begin{equation*}
            \Delta(s,a) \doteq r(s,a) + \gamma \E_{s' \sim{} p} V(s') - Q(s,a).
            \label{eq:exact-delta#1}
        \end{equation*}
    \label{thm:gen_double_sided_bounds#1}
    \end{customthm}
    
}

\def\RwdCor#1{
\begin{customcor}{1}
Consider an MDP with bounded continuous state-action space $\s \times \A \subset \mathbb{R}^d$, and deterministic dynamics. Let $\varepsilon~>~0,\ \delta~>~0$, and $\left|\mathcal{B}\right|~\geq~\left( \frac{L_r \mathrm{diam}(\s\times\A)}{\varepsilon} \right)^{d} \log 1/ \delta$, be given, where $\mathrm{diam}$ represents the diameter of a bounded space. Suppose $|\mathcal{B}|$ samples are drawn uniformly from state-action space, $(s,a) \sim{} \mathrm{Unif}\left(\s\times \A\right)$, and denote the convex hull of these points as $c\doteq \mathrm{Conv}(\mathcal{B})$. Then, the following bounds on the reward function's extrema
\begin{align*}
     \inf_c r(s,a) &\geq \min_{(s,a) \in \mathcal{B}} r(s,a) - \varepsilon, \\
    \sup_c r(s,a) &\leq \max_{(s,a) \in \mathcal{B}} r(s,a) + \varepsilon,
\end{align*}
hold with probability at least $1-\delta$.
\label{cor:reward-bounds#1}
\end{customcor}
}

\def\OnTheFlyBoundStd#1{
    \begin{customthm}{4.9}
        Consider a (standard RL) task with reward function $r(s,a)$ and (unknown) optimal value function $Q^*(s,a)$.
        Let an estimate for the state value function be given as $V(s)$.

        The optimal value function $Q^*(s,a)$ is then bounded by:
        
        \begin{align}
            Q^*(s,a) &\geq r(s,a) + \gamma \left(\E_{s' \sim{} p} V(s') + \frac{\inf \Delta}{1-\gamma}\right) \\
            Q^*(s,a) &\leq r(s,a) + \gamma \left(\E_{s' \sim{} p} V(s') + \frac{\sup\Delta}{1-\gamma}\right) \label{eq:std_gen_double_sided_bounds#1}
        \end{align}
        where         
        \begin{equation*}
            \Delta(s,a) \doteq r(s,a) + \gamma \E_{s' \sim{} p} V(s') - V(s).
        \end{equation*}
    \label{thm:std_gen_double_sided_bounds}
    \end{customthm}
    In Eq. \eqref{eq:std_gen_double_sided_bounds#1}, the $\inf$ and $\sup$ are taken over the continuous state-action space $\s \times \A$.
}

\def\TightOnTheFlyBound#1{
    \begin{customlemma}{4.1a}[]
        Consider an entropy-regularized MDP $\langle \s, \A, p, r, \gamma, \beta \rangle$ with (unknown) optimal value function $Q^*(s,a)$.
        Let an estimate for the value function $Q(s,a)$ be given.
        Denote $V(s)~\doteq~1/\beta \log \E_{a \sim \pi_0} \exp \beta Q(s,a)$. Suppose there exists an ``identity'' action $a_\emptyset(s) \in \A$ for each state, which deterministically transitions the agent to the same state: $p(s'|s,a_\emptyset(s) ) = \delta(s'-s)$ for all $s \in \s$.

        Then the lower bound on the optimal value function $Q^*(s,a)$ can be improved:
        
        \begin{equation}
            % Q^*(s,a) \geq Q(s,a) + \frac{\Delta(s,a)}{1-\gamma} %\leq  \leq r(s,a) + \gamma \left(\E_{s' \sim{} p} V(s') + \frac{\sup\Delta}{1-\gamma}\right) 
            Q^*(s,a) \geq r(s,a) + \gamma \left( V(s') + \frac{1}{1-\gamma}\Delta(s', a_\emptyset) \right)
            \label{eq:tighter_double_sided#1}
        \end{equation}
        % where         
        % \begin{equation*}
        %     \Delta(s,a) \doteq r(s,a) + \gamma \E_{s' \sim{} p} V(s') - Q(s,a).
        % \end{equation*}
    \end{customlemma}
    % In Eq. \eqref{eq:tighter_double_sided#1}, the $\sup$ is taken over all state-action pairs $(s,a) \in \s \times \A$.
}

\def\PolicyImprovementEqn#1{
    \begin{customthm}{5.6}
        Given a policy $\pi_k$ in entropy-regularized RL with soft value $Q^{\pi_k}(s,a)$, then the improved policy $\pi_{k+1} \propto \exp \beta Q^{\pi_k}(s,a)$ has a soft value, $Q^{\pi_{k+1}}(s,a)$ given by:
        \begin{equation}
            Q^{\pi_{k+1}}(s,a) = Q^{\pi_{k}}(s,a) + q^{\pi_{k+1}}(s,a)
        \end{equation}
        where $q^{\pi_{k+1}}(s,a)$ is the value for policy $\pi_{k+1}$ in standard RL for a task with reward function $r(s,a) = r(s,a) + \gamma \E_{s'}V^{\pi_k}(s') - Q^{\pi_k}(s,a) $ 
    \label{thm:PI_stdPE}
\end{customthm}
}

\def\BoundingContinuousExtrema#1{
    \begin{customlemma}{4.4}
        Let $\s \times \A$ be a bounded metric space with diameter $D$, %$d((s_1,a_1),(s_2,a_2))=d_{\s}(s_1,s_2) + d_{\A}(a_1,a_2)$
        and let $r: \s \times \A \to \mathbb{R}$ be $L_r$-Lipschitz (w.r.t. the same metric). %Then, given the diameters of $\s$ and $\A$ as $D_{\s}, D_{\A}$, respectively, one can bound 
        Then the global extrema of $r(s,a)$ on $\s \times \A$ are bounded as follows:
        \begin{align*}
            \sup_{s\in \s,a \in \A} r(s,a) &\leq \min_{(s,a) \in \mathcal{D}}  r(s,a) + L_r D  \\ 
            \inf_{s\in \s,a \in \A} r(s,a) & \geq \max_{(s,a) \in \mathcal{D}} r(s,a) - L_r D
        \end{align*}
        where $\mathcal{D}$ is the dataset of $(s,a)$ tuples available for querying the magnitude of $r$ (e.g. the current batch or buffer).
        \label{lem:extrema#1}
    \end{customlemma}
}

\def\VfuncBound#1{
\begin{customthm}{4.5}
% Let an entropy-regularized MDP be given with $L_r$-Lipschitz reward function and $L_p$-Lipschitz transition dynamics. Using a Gaussian parameterization for the policy with a lower-bounded variance, suppose that an estimate for $Q^\pi$ is known such that $|Q^\pi(s,a)-\bar{Q}^{\pi}(s,a)|\leq \varepsilon.$ Further, suppose $\gamma L_p (1+L_\mathcal{N})<1$, where $L_\mathcal{N} =\sigma_{\text{min}}^{-2}(2\pi e)^{-1/2}$ is the Lipschitz constant of the Gaussian distribution, $\mathcal{N}(\mu, \sigma)$. %As in SAC \cite{stable-baselines3}, let $\mu(s)$ and $\log p(a=\mu(s))$ be given.

Let an entropy-regularized MDP be given with an $L_Q$-Lipschitz value function $\bar{Q}^\pi$. Using a Gaussian parameterization for the associated policy $\pi(\cdot|s)=\mathcal{N}(\mu(s),\sigma(s))$, suppose that $\bar{Q}^\pi$ is an $\varepsilon$-optimal approximation of the policy's true value, $Q^\pi$.

% Let the expectation over $Q(s, \cdot)$ values in the calculation of $V^\pi(s)$
% be estimated by a single action sample (the mean of the Gaussian): $a \to \mu(s)$
By estimating the state-value function as:
% \begin{equation}
%     \bar{V}^\pi(s) = \bar{Q}^\pi(s,\mu) + \beta^{-1} \mathbb{H}\left[\pi(\cdot|s) \right]
% \end{equation}
\begin{equation}
    \bar{V}^\pi(s) = \bar{Q}^\pi(s,\mu) - \frac{1}{\beta} \E_{a\sim{} \pi} \log\frac{\pi(a|s)}{\pi_0(a|s)},
    \label{eq:approx_V#1}
\end{equation}
% \begin{equation}

the error in using such an approximation is upper bounded:
\begin{equation*}
    |\bar{V}^\pi(s) - V^\pi(s)| \leq \sqrt{\frac{2 }{\pi}}L_Q \sigma(s) e^{-\mu(s)^2/2\sigma(s)^2}  + \varepsilon
\end{equation*}
% with 
In the case that the function $Q$ used is an optimal value function for an $(L_r, L_p)$-Lipschitz task, with a policy whose variance is lower bounded $\sigma(s) \geq \sigma_\text{min}$ and $\gamma L_p (1+L_\mathcal{N})<1$, where $L_\mathcal{N} =\sigma_{\text{min}}^{-2}(2\pi e)^{-1/2}$ is the Lipschitz constant of the Gaussian distribution, then the Lipschitz constant for $Q$ can be computed as:
% \JA{the error $\varepsilon$ is because the gaussian-param'd policy does not satisfy policy evaluation equation in exact case}
\begin{equation}
    L_Q=\frac{L_r + \gamma L_p (\beta \sigma_{\min})^{-1}}{1-\gamma L_p (1+L_\mathcal{N})}.
    \label{eq:L_Q_entreg#1}
\end{equation}
%  is known such that $|Q^\pi(s,a)-\bar{Q}^{\pi}(s,a)|\leq \varepsilon.$ Further, suppose , $\mathcal{N}(\mu, \sigma)$. %As in SAC \cite{stable-baselines3}, let $\mu(s)$ and $\log p(a=\mu(s))$ be given.

%     \bar{\Delta}(s,a) = r(s,a) + \gamma Q(s',\mu) + \gamma \alpha H\left[ \pi(\cdot|s')\right] - Q(s,a)
% \end{equation} be the one-point estimate of $\Delta$ where the expectation has been replaced by the mean action, and $s' \sim{} p(\cdot | s,a)$ is the next action. Let $\pi(\cdot |s')$ be parameterized by a Gaussian distribution $\mathcal{N}(\mu, \sigma)$. Then, the difference between the exact reward function ($\Delta$) and the estimate ($\bar{\Delta}$) can be bounded as:

% \begin{equation}
%     |\Delta(s,a) - \bar{\Delta}(s,a)| \leq \frac{\gamma L_r}{1-\gamma L_p} \left( \sqrt{\frac{2 }{\pi}}\sigma + \frac{e^{-1/2\sigma^2}}{\pi} \right)
% \end{equation}
\label{thm:vfunc_bound#1}
\end{customthm}
}
\def\Bogoliubov#1{
    \begin{customthm}{5.3}
        % \begin{equation}
        %     K^*(s,a) \leq \E_{\tau \sim{} \pi_f} \sum_{t=0}^{\infty} \gamma^t \kappa(s_t, a_t)
        % \end{equation}
        % \begin{equation}
        %     \widetilde{Q}^*(s,a) \leq S_\tau(\pi|\pi_0) + \E_\pi \sum_{t=0}^{\infty} \gamma^t \widetilde{r}(s_t, a_t)
        % \end{equation}
        \begin{equation}
            \widetilde{Q}^*(s,a) \geq \E_\pi \sum_{t=0}^{\infty} \gamma^t \left( \widetilde{r}(s_t, a_t) + \frac{1}{\beta} \textrm{KL}(\pi|\pi_0) \right)
        \end{equation}
        \label{lem:theorem}
    \end{customthm}

}

\def\ClippedBOperatorThm#1{
\begin{customprop}{1}
Let $B(\cdot)$ denote the Bellman operator, and let the functions $L(s,a),\ U(s,a)$ be lower and upper bounds on the optimal action-value function respectively: $L(s,a)~\leq~Q^*(s,a)~\leq U(s,a)$ for all $s \in \s$ and $a\in\A$. The clipped Bellman operator, $B_{C}Q(s,a)~:=~\max \left( \min \left(BQ(s,a), U(s,a) \right), L(s,a) \right)$ converges to the optimal action-value function $Q^*(s,a)~=~B^\infty Q(s,a)=~B_C^\infty Q(s,a)$ for any bounded initial function $Q(s,a)$. 
% \JA{The operator $\mathcal{B}_C$ enjoys a contraction rate of $\gamma' \leq \gamma$.}
\end{customprop}
}

\def\vanNiekerkExtension#1{
\begin{customthm}{5.1}
    Consider $m$ solved tasks in the entropy-regularized setting, with reward functions $\{r_1,\dotsc,r_m\}$ varying only on the set of absorbing states. Assume all tasks are given with the same deterministic dynamics. Given a set  of non-negative weights $w_j$, consider a new task with the same reward function for the interior (i.e. non-absorbing) states and with reward function for the absorbing states given by
    \begin{equation}
        \widetilde{r}(s,a) = \tau \log \sum_{j=1}^m w_j e^{r_j(s,a) / \tau}.
    \end{equation}
    Then, the optimal value function for such a task is given by:
    \begin{equation}
    \widetilde{Q}(s,a) = \tau \log \sum_{j=1}^m w_j e^{Q_j(s,a) / \tau}.
    \end{equation}
    % for any $\textbf{w}\in \mathbb{R}^m$ such that $w_j > 0$.
    \label{thm:niek}
\end{customthm}
}

\def\FullPACInformal#1{
\begin{customthm}{2 (Informal)}
Consider an MDP with a bounded continuous state and action space, $\s~\times~\A~\subset~\mathbb{R}^d$, with stochastic dynamics. Suppose an $L_Q$-Lipschitz function $Q(s,a)$ is given to generate double-sided bounds on the optimal value function, denoted $Q^*(s,a)$. Let $\varepsilon_i>~0, \delta_i>~0$ be given and define the horizon $H=(1-\gamma)^{-1}$, and sample budgets: $\left|\mathcal{B}\right|~\geq~\mathcal{O}\left(\varepsilon_1^{-d} \log \delta_1^{-1}\right),\ n_\s \geq \mathcal{O}\left(H^2 \varepsilon_2^{-2} \log \delta_2^{-1}\right),\ n_\A \geq \mathcal{O}\left( e^{\beta (H + \varepsilon_3)} \log \delta_3^{-1}\right).$
% \begin{align}
%     \left|\mathcal{B}\right|~&\geq~\mathcal{O}\left(\varepsilon_1^{-d} \log \delta_1^{-1}\right),\\ %\left( \frac{L_\Delta \mathrm{diam}\left(\s \times \A \right)}{\varepsilon_1} \right)^{d} \log \frac{1}{\delta_1}, \\
%     n_\s &\geq \mathcal{O}\left(H^2 \varepsilon_2^{-2} \log \delta_2^{-1}\right),n_\s &\geq \mathcal{O}\left(H^2 \varepsilon_2^{-2} \log \delta_2^{-1}\right),\  \\% \frac{1}{2}\left( \frac{H(R_\mathrm{max}-R_\mathrm{min})}{\varepsilon_2}\right)^2 \log \frac{2}{\delta_2}, \\
%     n_\A &\geq \mathcal{O}\left( e^{\beta (H + \varepsilon_3)} \log \delta_3^{-1}\right). %\frac{1}{2} \log\frac{2}{\delta_3} \left( \frac{e^{\beta H (R_{\mathrm{max}}-R_{\mathrm{min}})}-1}{e^{-\beta \varepsilon_3} -1 } \right)^2.
% \end{align}
%Suppose $|\mathcal{B}|$ samples are drawn uniformly from the joint state-action space, {$(s,a)~\sim{}~\mathrm{Unif}\left(\s~\times\A\right)$} to estimate the extrema of $\hat{\Delta}$. 
Suppose $n_\s$ samples are used to estimate the expectation over next-states and $n_\A$ samples are used to estimate the expectation over next-actions in the soft state-value function. %via Eq.~\eqref{eq:V-approx}. 
% Then, for $\hat{V},\ \hat{\Delta}$ given in Equation~\eqref{eq:V-approx}, \eqref{eq:deltahat-defn}, 
Denoting $\hat{V}, \hat{\Delta}$ as the quantities estimated from samples, the following bounds% on the $Q$-values
{ %\small
\begin{align}
    Q^*(s,a) &\leq r(s,a)+\gamma \biggr(\frac{1}{n_\s}\sum_{i=1}^{n_\s}\hat{V}(s'_i) + \frac{\max_{(s,a) \in \mathcal{B}} \hat{\Delta}(s,a) + \varepsilon_1 + \varepsilon_2+ \varepsilon_3   }{1-\gamma} \biggr) \\
    Q^*(s,a) &\geq  r(s,a)+\gamma \biggr(\frac{1}{n_\s}\sum_{i=1}^{n_\s}\hat{V}(s'_i) + \frac{\min_{(s,a) \in \mathcal{B}} \hat{\Delta}(s,a) - \varepsilon_1 - \varepsilon_2- \varepsilon_3}{1-\gamma} \biggr)
\end{align}
}
hold with probability at least $1-\delta_1 - 2\delta_2 - 2\delta_3$. 
\label{thm:final-prop#1}
\end{customthm}
}

\def\FullPACInformalSimple#1{
\begin{customthm}{2 (Informal)}
Consider an MDP with a bounded continuous state and action space, $\s~\times~\A~\subset~\mathbb{R}^d$, with stochastic dynamics. Suppose an $L_Q$-Lipschitz function $Q(s,a)$ is given to generate double-sided bounds on the optimal value function, denoted $Q^*(s,a)$. Let $\varepsilon>~0, \delta>~0$ be given and define the horizon $H=(1-\gamma)^{-1}$, and sample budgets: $\left|\mathcal{B}\right|~\geq~\mathcal{O}\left(\varepsilon^{-d} \log \delta^{-1}\right),\ n_\s \geq \mathcal{O}\left(H^2 \varepsilon^{-2} \log \delta^{-1}\right),\ n_\A \geq \mathcal{O}\left( e^{2\beta (H - \varepsilon)} \log \delta^{-1}\right).$

Suppose $n_\s$ samples are used to estimate the expectation over next-states and $n_\A$ samples are used to estimate the expectation over next-actions in the soft state-value function. 
Denoting $\hat{V}, \hat{\Delta}$ as the quantities estimated from samples, the following bounds% on the $Q$-values
{ %\small
\begin{align}
    Q^*(s,a) &\leq r(s,a)+\gamma \biggr(\frac{1}{n_\s}\sum_{i=1}^{n_\s}\hat{V}(s'_i) + \frac{\max_{(s,a) \in \mathcal{B}} \hat{\Delta}(s,a) + \varepsilon }{1-\gamma} \biggr) \\
    Q^*(s,a) &\geq  r(s,a)+\gamma \biggr(\frac{1}{n_\s}\sum_{i=1}^{n_\s}\hat{V}(s'_i) + \frac{\min_{(s,a) \in \mathcal{B}} \hat{\Delta}(s,a) - \varepsilon}{1-\gamma} \biggr)
\end{align}
}
hold with probability at least $1-\delta$. 
\label{thm:final-prop#1}
\end{customthm}
}

\def\FullPAC#1{
\begin{theorem}
Consider an MDP with a bounded continuous state and action space, $\s~\times~\A~\subset~\mathbb{R}^d$, with stochastic dynamics. Suppose an $L_Q$-Lipschitz function $Q(s,a)$ is given to generate double-sided bounds on the optimal value function, denoted $Q^*(s,a)$. Let $\varepsilon_i>~0, \delta_i>~0$ be given and define the horizon $H=(1-\gamma)^{-1}$, and sample budgets
\begin{align}
    \left|\mathcal{B}\right|~&\geq~\left( \frac{L_\Delta \mathrm{diam}\left(\s \times \A \right)}{\varepsilon_1} \right)^{d} \log \frac{1}{\delta_1}, \\
    n_\s &\geq  \frac{1}{2}\left( \frac{H(R_\mathrm{max}-R_\mathrm{min})}{\varepsilon_2}\right)^2 \log \frac{2}{\delta_2}, \\
    n_\A &\geq \frac{1}{2}   \left( \frac{e^{\beta H (R_{\mathrm{max}}-R_{\mathrm{min}})}-1}{e^{\beta \varepsilon_3} -1 } \right)^2 \log\frac{2}{\delta_3}.
\end{align}

Suppose $|\mathcal{B}|$ samples are drawn uniformly from the state-action space, {$s~\sim{}~\mathrm{Unif}\left(\s\right)$} and {$a~\sim{}~\mathrm{Unif}\left(\A\right)$} to estimate the extrema of $\hat{\Delta}$. Suppose $n_\s$ samples are used to estimate the expectation over next-states and $n_\A$ samples are used to estimate the soft state-value function %via Eq.~\eqref{eq:V-approx}. 
% Then, for $\hat{V},\ \hat{\Delta}$ given in Equation~\eqref{eq:V-approx}, \eqref{eq:deltahat-defn}, 
Denoting $\hat{V}, \hat{\Delta}$ as the quantities estimated from samples, the following bounds on the $Q$-values
{ %\small
\begin{align}
    Q^*(s,a) &\leq r(s,a)+\gamma \biggr(\frac{1}{n_\s}\sum_{i=1}^{n_\s}\hat{V}(s') + \frac{\max_{(s,a) \in \mathcal{B}} \hat{\Delta}(s,a) + \varepsilon_1 + \varepsilon_2+ \varepsilon_3   }{1-\gamma} \biggr) \\
    Q^*(s,a) &\geq  r(s,a)+\gamma \biggr(\frac{1}{n_\s}\sum_{i=1}^{n_\s}\hat{V}(s') + \frac{\min_{(s,a) \in \mathcal{B}} \hat{\Delta}(s,a) - \varepsilon_1 - \varepsilon_2- \varepsilon_3}{1-\gamma} \biggr) 
\end{align}
}
hold with probability at least $1-\delta_1 - 2\delta_2 - 2\delta_3$. 
\label{thm:final-prop-formal#1}
\end{theorem}

}
\def\FullyPropagated#1{
\begin{customthm}{4.6}
% Let the optimal value function for some Lipschitz task be denoted $q^*$, for which an $\varepsilon$-optimal approximation, $q^\pi$ is given such that $|q^\pi(s,a)-q^*(s,a)| \leq \varepsilon$ for all $s,a$.
Let the $L_Q$-Lipschitz value function $Q^\pi$ and corresponding Gaussian policy $\pi(\cdot|s) = \mathcal{N}(\mu(s), \sigma(s))$ be given, where $Q^\pi$ is an $\varepsilon$-optimal estimate of the true policy's value function.
For an $(L_r, L_p)$-Lipschitz task with (unknown) optimal value function $Q^*$, let $\bar{V}^\pi$ be the one-point estimate of the (known) value function $Q^\pi$, and denote $\bar{\Delta}(s,a) = r(s,a) + \gamma \E_{s' \sim{} p} \bar{V}^\pi(s') - Q^\pi(s,a)$. Then:
\begin{align*}
    % Q^*(s,a) &\leq r(s,a) + \gamma \E_{s' \sim{} p} \bar{v}^\pi(s') + \\ 
    % & \frac{\gamma}{1-\gamma} \left[\min_{(s,a)\in \mathcal{D}} \bar{\Delta}(s,a) + L_\Delta ||(D_{\s},D_{\A})||_p + A +2 \varepsilon \right]  \\
    % Q^*(s,a) &\leq r(s,a) + \gamma \left(\E_{s' \sim{} p}\bar{V}^\pi(s') + \E_{s'\sim{} p}A(s')
    &Q^*(s,a) \leq r(s,a) + \gamma \E_{s' \sim{} p}\left[\bar{V}^\pi(s') + A(s')\right]      % &+\frac{1}{1-\gamma}\left(\min_{(s,a) \in \mathcal{D}} \left(\bar{\Delta}(s,a) + \gamma A + (1+\gamma)\varepsilon \right) + L_\Delta D  \right)\right) \\
    +\frac{\gamma}{1-\gamma}\left(\min_{(s,a) \in \mathcal{D}} \left(\bar{\Delta}(s,a) + \gamma \E_{s'\sim{} p} A(s') \right) + L_\Delta D  \right) \\% &+\frac{1}{1-\gamma}\left(\min_{(s,a) \in \mathcal{D}} \left(\bar{\Delta}(s,a) + \gamma A + (1+\gamma)\varepsilon \right) + L_\Delta D  \right)\right) \\
    % +\frac{1}{1-\gamma}\left(\min_{(s,a) \in \mathcal{D}} \left(\bar{\Delta}(s,a) + \gamma A(s) \right) + L_\Delta D  \right)\right) \\
    % Q^*(s,a) &\geq r(s,a) + \gamma \E_{s' \sim{} p} \bar{v}^\pi(s') + \\ &\frac{1}{1-\gamma} \left[\max_{(s,a)\in \mathcal{D}} \bar{\Delta}(s,a) - L_\Delta ||(D_{\s},D_{\A})||_p - A - 2\varepsilon \right] 
    &Q^*(s,a) \geq r(s,a) + \gamma \E_{s' \sim{} p}\left[\bar{V}^\pi(s') - A(s')\right]     % &+\frac{1}{1-\gamma}\left(\min_{(s,a) \in \mathcal{D}} \left(\bar{\Delta}(s,a) + \gamma A + (1+\gamma)\varepsilon \right) + L_\Delta D  \right)\right) \\
    +\frac{\gamma}{1-\gamma}\left(\max_{(s,a) \in \mathcal{D}} \left(\bar{\Delta}(s,a) - \gamma \E_{s'\sim{} p}A(s') \right) - L_\Delta D  \right) \\%
\end{align*}
where for brevity we have introduced the notations  \begin{equation}
    A(s)= \sqrt{\frac{2}{\pi}}L_Q\sigma(s) e^{-\mu(s)/2\sigma(s)^2}+\varepsilon
\end{equation} and 
%$L_\Delta~=~\max \left\{L_Q, \gamma \left(L_Q(1+L_\mathcal{N}) + (\beta \sigma_{\text{min}})^{-1}\right)\right\}$ 
\begin{equation}
    L_\Delta=\max \left\{L_r, L_Q, \gamma L_p \left(L_Q(1+L_\mathcal{N}) + (\beta \sigma_{\text{min}})^{-1}\right)\right\}
\end{equation} with $D$ %~=~||(D_\mathcal{S},D_{\mathcal{A}})||_p$ 
denoting the diameter of the state-action space.
%in the $p$-norm.
%$L_\Delta=\frac{ L_r}{1-\gamma L_p(1+L_\mathcal{N})}$.
\label{thm:fully_propagated#1}
\end{customthm}
}

\maketitle
\begin{abstract}
    An agent’s ability to leverage past experience is critical for efficiently solving new tasks.
    Prior work has focused on using value function estimates to obtain zero-shot approximations for solutions to a new task. In soft $Q$-learning, we show how any value function estimate can also be used to derive double-sided bounds on the optimal value function. The derived bounds lead to new approaches for boosting training performance which we validate experimentally. Notably, we find that the proposed framework suggests an alternative method for updating the $Q$-function, leading to boosted performance. 
\end{abstract}
% \JA{TODOs:

% \begin{itemize}
%     \item check citation guideline in instructions doc
%     \item PAC bound?
% \end{itemize}

% }
\section{Introduction}
% \JA{scrap -- too generic for RLC, make better case for prior knowledge (usually available but typically thrown away}
In recent years, reinforcement learning (RL) has seen impressive success at the price of ever-increasing sample budgets. The current paradigm of RL consists of training agents from scratch with new hyperparameters or in new domains, without significant reuse of previously collected information. The large datasets generated from such runs have been approached with techniques such as offline RL; however, the approximate solutions obtained from previous runs are typically not reused. To address this issue, approaches that directly leverage this learned prior knowledge to efficiently calculate policies for new tasks are needed. While prior solutions may not be \textit{optimal} for arbitrary new tasks, they have been shown to serve as useful approximations that reduce training time in a variety of settings: \citep{rusu2016progressive, tasse2021generalisation, agarwal2022reincarnating, Adamczyk_UAI}. In this work, we present a new way in which information from previous solutions can be further leveraged
%the use of approximations based on prior knowledge 
to address new tasks\footnote{Our code is publicly available at \url{https://github.com/JacobHA/RLC-SoftQBounding}.}. %\JA{the way this is currently phrased, sounds like it is already a solved problem!}
% \JA{need solutions that harness prior knowledge}

% \JA{What current solutions do}
%If agents are trained to learn the action-values $Q(s,a)$, further information can be extracted from known solutions. In particular, we show that bounds on the optimal value function (for any new task of interest) can be calculated from any previous task's $Q$-values.
%\JA{this is not a good intro sentence. maybe better somewhere else/prior work}
Previous work has focused on addressing this challenge with approaches such as transfer learning, curriculum learning, and compositionality. In this work, we will focus on value-based RL algorithms where the agent learns the optimal action-value, or $Q$-function. 
In many instances, the agent has an estimate of the value function even before training begins. For example, in the case of curriculum learning, the agent has the $Q$-values for previously learned (progressively more challenging) tasks. In the case of compositional \citep{Haarnoja2018} or hierarchical RL \citep{hafner2022deep}, the agent can combine knowledge by applying a function on the subtasks' $Q$-values. When using an exploratory skill-acquisition approach~\citep{eysenbach2018diversity} or constructing a task basis~\citep{alver2021constructing}, the agent obtains solutions for a diverse set of skills to use on downstream tasks.
Even in cases where an initial estimate is not explicitly provided, the agent indeed has access to an estimate through the $Q$-values obtained in the ongoing learning phase (bootstrapping). %\JA{how to connect / prompt question in readers head?} However, besides using them as a warmstart, it is not clear how to employ these prior (or ongoing) estimates of $Q$-functions to aid the solution of the task at hand.

An underlying question in these scenarios is the following: How can the agent use these known value function estimate(s) for solving a new target task? Does the estimate only serve as a zero-shot approximation or is there additional useful information that can be extracted from such estimates? {\textbf{In this work, we address this question by deriving bounds on the \textit{optimal} $Q$-values from an \textit{arbitrary} prior estimate.}} We emphasize that the surprising nature of these bounds is that information concerning the \textit{optimal} value function can be derived from arbitrarily \textit{suboptimal} estimates.

% \JA{what is the connection?}
To derive such bounds, we leverage exact results on the $Q$-function from recent work by \cite{cao2021identifiability} and \cite{Adamczyk_AAAI}. In the latter, the authors show (in their Theorem 1) that there exists a method of ``closing the gap'' between any estimate (therein denoted $Q^*(s,a)$) and any target (denoted $\widetilde{Q}^*(s,a)$) task in entropy-regularized RL.
%This statement is facilitated by \cite{cao2021identifiability}'s \CaoThmNospace{}, which can be 
Here, we show that since this gap between the target and estimated value functions, $\widetilde{Q}^*(s,a)-Q^*(s,a)~=~K^*(s,a)$, is itself an optimal value function, it can be bounded. As a consequence, instead of providing only a zero-shot approximation (``warmstart'' or ``jumpstart'') for training the target task, we show that the estimates available to the agent also provide a double-sided bound on the desired optimal $Q$-values. From Theorem 1 of \cite{cao2021identifiability}, it can be shown that an optimal solution is not required for deriving such bounds, and in fact \textit{any} function over the state-action space can be used to derive double-sided bounds, including for instance the bootstrapped estimate of $Q(s,a)$. Since it is the most general, we focus on this case in the present work.

\begin{figure}[t]
    \centering
    \includegraphics[width=0.75\textwidth]{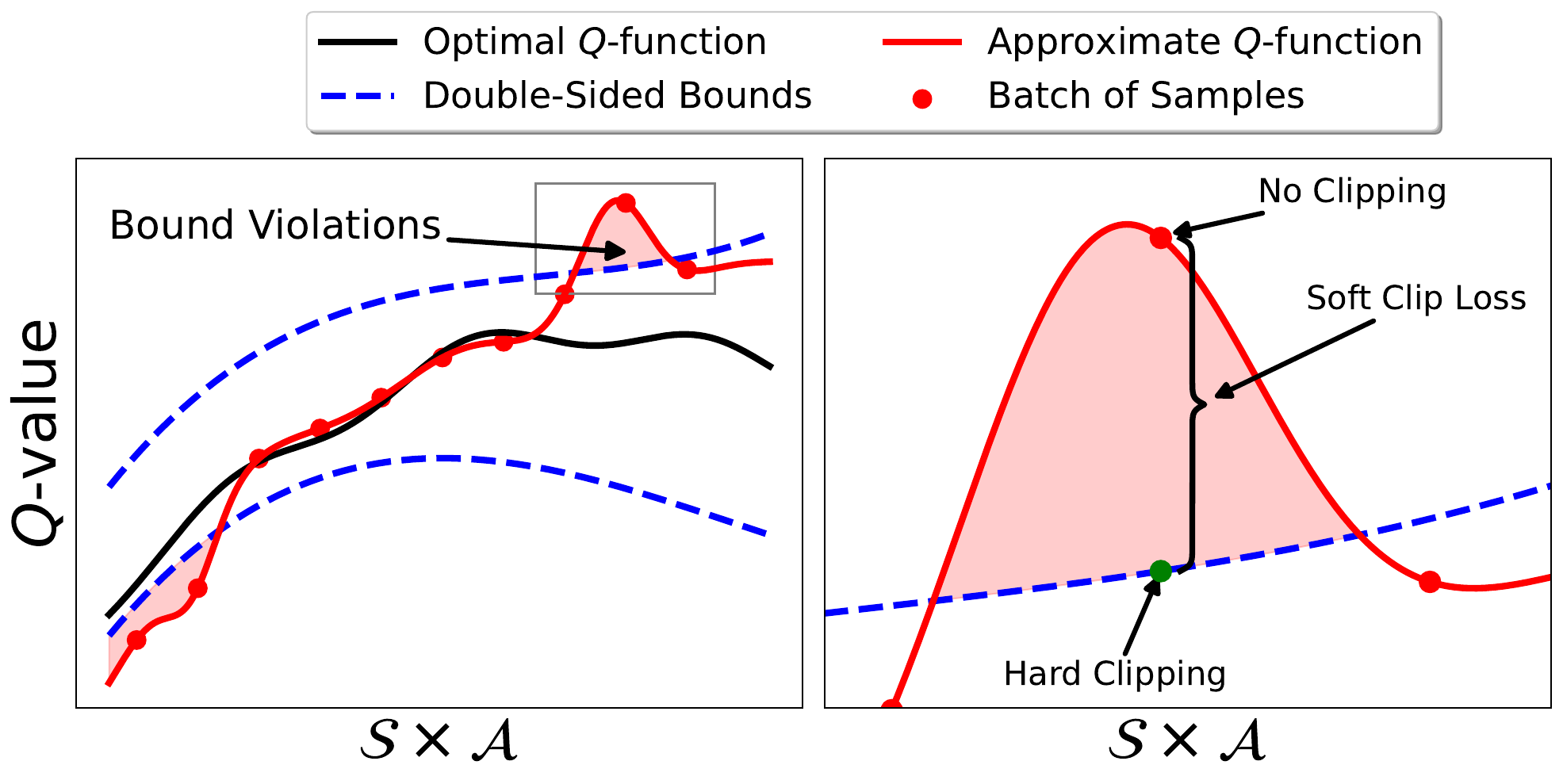}
    \caption{Schematic illustration of the main contribution of this work. Given any approximation (red curve) to the optimal value function of interest (black curve), we derive double-sided bounds (blue curves) that lead to clipping approaches during training. Based solely on the current approximation for $Q(s,a)$ (red curve), we derive double-sided bounds on the \textit{unknown} optimal value function $Q^*(s,a)$ (black curve). In the right panel, we show the different clipping methods, which are described further in the ``Experimental Validation'' section. In ``Hard Clipping'', the target is replaced with the exceeded bound; in ``Soft Clipping'', an additional loss term is appended to the Bellman loss, proportional to the magnitude of the bound violation.}
    \label{fig:schematic}
\end{figure}

A schematic illustration of our approach is provided in Figure~\ref{fig:schematic}. 
Starting with samples of a value function (red points), we derive double-sided bounds (dashed blue lines) on the optimal value function (solid black line).
We find that applying these bounds during training significantly boosts the agent's training performance in the tabular setting. 
We provide further theoretical analysis in continuous state-action spaces, relevant for the function approximator (FA) setting in deep reinforcement learning, for which we present initial experiments in Section~\ref{sec:experiments}.

\textbf{Main contributions}
\newline
The main contributions of our work are as follows:
\begin{enumerate}
    % \item Convert any function $f: \s \times \A \to \mathbb{R}$ (typically a value function estimate) to a double-sided bound on the optimal value function for any task with the same state-action space, $Q^*(s,a)$.
    \item A framework for bounding optimal value functions given any estimate of the value function.
    \item A novel soft $Q$-learning algorithm and demonstration of its advantages.
    \item Extension of theoretical results to continuous state-action spaces. 
\end{enumerate}
% \JA{change ``convex combinations'' to ``specific classes of functions'' (e.g. boolean) and change ``general'' to ``any''?}

% There are multiple applications that arise from the derivation of such double-sided bounds. The bounds (1) allow confinement of FA training to a limited output range, (2) provide a mechanism to choose the ``best'' skill from a pre-trained set of skills and (3) establish a framework that provides insights into and extends previous results on exact compositions of value functions.
% \JA{ RM?:
% We will highlight two applications that arise from the derivation of such double-sided bounds. The bounds (1) allow confinement of FA training to a limited output range and (2) provide a mechanism to choose the ``best'' skill from a pre-trained set of skills. \JA{do this in tabular!!!}
% }
%\JA{ What problem(s) does this solve? Our work can address them ... which we demonstrate in the experiments. \textbf{Add more motivation}}

% \JA{Begs the question of how good is the bound given it comes from an arbitrary function. Explain: Can prove this bound becomes tight as the input becomes cloesr to optimal value function (i.e. $\Delta \to 0$). }

\section{Preliminaries}
For the theoretical analysis, we begin with finite, discrete state and action spaces, and we subsequently extend our analysis to continuous spaces. In this setting, the RL problem is modeled as a Markov Decision Process (MDP) represented by the tuple $\langle \s,\A,p,r,\gamma \rangle$ where $\s$ is the state space; $\A$ is the action space; $p: \s \times \A \to \s$ is the transition function (dynamics); $r: \s \times \A \to \mathbb{R}$ is a (bounded) reward function; and $\gamma \in [0,1)$ is the discount factor.
% The objective in standard (un-regularized) RL is to find an optimal policy that maximizes expected rewards collected by the agent, i.e.
% \begin{equation}\label{eq:std_rl_obj}
%     \pi^* = \arg\max_{\pi} \mathbb{E}
%     %_{\rho_\pi} 
%     \left[ \sum_{t=0}^{\infty} \gamma^{t} r(s_t,a_t) \right].
% \end{equation}
We focus on the generalization of entropy-regularized RL \citep{ZiebartThesis}, which augments the un-regularized RL objective by including an entropic regularization term which penalizes control over a pre-specified
reference policy:
\begin{equation*}%\label{eq:optimal_policy_2}
    \pi^* = \arg\max_{\pi}
    \mathbb{E}
    %_{\rho_\pi}%_{\pi(a|s),p(s'|s,a)}
    \left[ \sum_{t=0}^{\infty} \gamma^{t} \left( r_t - \frac{1}{\beta}
        % \pi^0(a_t|s_t)
        \log\left(\frac{\pi(a_t|s_t)}{\pi_0(a_t|s_t)} \right) \right) \right]
\end{equation*}
where $\pi_0(a|s)$ is a fixed prior policy.
This additional control cost discourages the agent from choosing policies that deviate too much from the prior policy.
Importantly, entropy-regularized MDPs lead to stochastic optimal policies that are provably robust to perturbations of rewards and dynamics \citep{eysenbach}; making for a more suitable approach to real-world problems.

The solution to the RL problem is defined by its optimal action-value function ($Q^*(s,a)$) from which one can derive the aforementioned optimal policy $\pi^*(a|s)$ through a Boltzmann distribution with temperature $\beta^{-1}$. The optimal value function can be obtained by iterating the following recursive Bellman equation
% In un-regularized RL, the Bellman optimality equation is given by \cite{suttonBook}:
% \begin{equation}
%     Q^*(s,a) = r(s,a) + \gamma \mathbb{E}_{s' \sim{} p(\cdot|s,a)} \max_{a'} \left( Q^*(s',a') \right).
%     \label{eq:bellman}
% \end{equation}
% \sout{ \ST{Eq.1 is defined for any policy $\pi^0$ might arise questions}} \JA{Am I missing something? I am looking at S+B's eq 4.2}
% The entropy term in the objective function of entropy-regularized RL modifies the previous optimality equation in the following way
\citep{ZiebartThesis, Haarnoja_SAC}:
\begin{equation}
    Q^*(s,a) = r(s,a) + \frac{\gamma}{\beta} \E_{s' \sim{} p
    %p(s'|s,a)} 
    }\log \E_{a' \sim{} \pi_0}
    %\sim{} \pi^0(a'|s')} 
    e^{ \beta Q^*(s',a') }.
    \label{eq:soft_bellman}
\end{equation}

The regularization parameter $\beta$ is used to control the degree of stochasticity in the optimal policy. In the entropy-regularized setting, $Q^*$ is referred to as the optimal ``soft'' action-value function. For brevity, we refer to $Q^*$ simply as the value function when context is clear.

\section{Prior Work\label{sec:prior}}
The use of value function bounds has been investigated in various domains of RL: offline and online settings, compositionality, and imitation learning. In this section, we briefly outline some of the most relevant work from this domain. 
%  in \cite{achiam2019towards}, bounds on the $Q$-function can help to avoid numerical divergences.}
%The importance of double-sided bounds on value functions has been explored in prior work. 
%In this section we review a set of the most relevant prior works: \cite{Nemecek,constrainedGPI,Haarnoja2018,Adamczyk_UAI,lee2021sunrise}. 
We contrast the existing literature with regard to the following features:
i) the MDP's structural assumptions,
ii) the requirement for additional samples to derive bounds,
iii) use of double or single-sided bounds.
% Specifically in the context of successor features, 

In \citep{Nemecek}, the authors have derived double-sided bounds on the state value function $V(s)$ when the task's reward function can be written as the \textit{positive conical combination} of subtask rewards. This method  requires additional samples for first learning the {\it successor features} (SFs) before then deriving the double-sided bounds for a downstream task. %The applicability of \cite{Nemecek} is limited to un-regularized RL.  
The aforementioned work was subsequently extended by \cite{constrainedGPI}, where, in the same SF setting, they present double-sided bounds on $Q$-values for linear combinations of subtask reward functions. They introduced a notion of ``soft clipping'' but it was not demonstrated in practice. We later adapt this idea of soft clipping to our setting, with details in Section~\ref{sec:experiments}. %As in the work of \cite{Nemecek}, the method by \cite{constrainedGPI} requires first learning {\it successor features} and is studied in the un-regularized RL formulation. 
Both \cite{Nemecek} and \cite{constrainedGPI} consider the un-regularized RL problem formulation ($\beta \to \infty$).

The two previous methods focus on the standard (un-regularized) reinforcement learning setting. However, the double-sided bounds presented by \cite{Haarnoja2018}'s Lemma 1 are derived for the MaxEnt setting, for the case of convex reward combinations. It is worth noting that the lower bound in this case must be learned (their $C$ function). Extending these results to other more general classes of functional composition, \cite{Adamczyk_UAI} provides double-sided bounds for both entropy-regularized and un-regularized RL. However, one side of the bounds in all cases must be learned as well.

% Finally, multiple prior works have focused on specific examples of compositionality  for which exact results can be obtained for the optimal value function.
% %In our framework, this corresponds to conditions for which the upper bound is identical to the lower bound. 
% These results typically involve multiple limiting assumptions on the structure of rewards functions, nature of transition dynamics and specific forms for the composition function. \cite{Todorov,vanNiekerk,boolean}.
In a different context focused on the stability of value-based RL, \cite{lee2021sunrise} proposes  
to (approximately) bound Bellman updates through a weighted ensemble, which improves the stability of training and sample efficiency in entropy-regularized RL. However, the method by \cite{lee2021sunrise} cannot leverage known solutions for new tasks, instead using a parallel ensemble of learners for variance estimation, exploiting the UCB framework~\citep{ucb} for exploration bonuses. 

Other examples of deep RL utilizing bounds include \citep{he2017learning}, which utilize bounds based on $n$-step returns, resulting in faster reward accumulation in the Atari suite. However, their bounds were not tested in stochastic environments but were shown to hold in expectation. Further, their upper bound depends on $Q^*$, the unknown optimal value function, making it intractable without first solving the task in question.
Later, the work of \cite{hoppe2020qgraph} modeled $Q$-functions through graphical models, using this structure to derive various bounds used in a constrained-DDPG algorithm. In this algorithm, a ``hard clipping'' mechanism is used, wherein the updates to the $Q$-values were clipped based on their bounds. We will consider a similar hard clipping approach, discussed in Section~\ref{sec:experiments}, but we derive bounds without imposing a graphical model of the dynamics. 
%-- should I elaborate here on stochastic dynamics and reducible vs irreducible?}.
%% I think it might be confusing at this stage and it isn't necessary
In principle, our bounds can interface easily with such prior work, by straightforwardly combining methods to obtain the tightest bounds possible.

% \JA{One reviewer thought this should be moved to the introduction section:}
In contrast to the aforementioned work, we propose a novel method for calculating double-sided bounds, not limited to a particular type of composition of prior solution(s) and valid for an \textit{arbitrary} input function. Our method for deriving double-sided bounds is zero-shot --- it does not require additional samples beyond those collected by the learning agent. Furthermore, our results are applicable to stochastic environments and both discrete or continuous domains. %We highlight the use of our theory in tabular domains and classic control environments. %The theoretical results are provided in the following ``Results'' section, and in the ``Applications'' section we demonstrate the applications of the theory in simple domains, leaving large scale experiments to future work. 

\section{Results}\label{sec:Theory}
%\JA{add emphasis on using entropy reg rather than un-regularized RL}
% In this section, we focus on entropy-regularized (MaxEnt) RL, the case considered in \cite{Adamczyk_AAAI}. The analogous results for un-regularized RL (which can be considered as a limiting case of entropy-regularized RL) are provided later. The proofs of all results shown can be found in the Appendix.
Our main result provides double-sided bounds on the optimal $Q$-function. We emphasize that \textit{any} (bounded) function $Q: \s \times \A \to \mathbb{R}$ can be used to generate such bounds. We suggestively use the notation ``$Q(s,a)$'' for this otherwise arbitrary function to emphasize that it may be derived from a previous task's solution, an estimate, or other ansatz (e.g. composition or hierarchical function) of $Q$-values.
\begin{tcolorbox}[colback=blue!5!white,colframe=blue!75!black]
\OnTheFlyBound{}
\end{tcolorbox}
In Equation~\ref{eq:gen_double_sided_bound}, the $\inf$ and $\sup$ are taken over the (potentially continuous) state-action space $\s \times \A$. The proof of Theorem~\ref{thm:gen_double_sided_bounds} can be found in Appendix~\ref{app:exact}. Intuitively, this result can be understood as a double-sided bound on the \textit{optimal} $Q$-function, calculated through a single iterate of the Bellman operator ($B$) on \textit{any} input function (denoted $Q$):
\begin{equation}
    \big|Q^*(s,a) - B Q(s,a)\big| \leq \mathcal{O}\left(H{\sqrt{\mathcal{L}}} \right),
\end{equation}
where $H=(1-\gamma)^{-1}$ denotes the effective time horizon and $\mathcal{L}$ denotes the Bellman loss incurred by the input function $Q$. During training, the Bellman loss (ideally) reduces to zero: $\mathcal{L}=||\Delta||^2 \to 0$, implying that $\inf \Delta \to 0$ and $\sup \Delta \to 0$, hence the bounds in Theorem~\ref{thm:gen_double_sided_bounds} are tight upon convergence of the soft action-value function. %Note that the loss can only become identically zero when the learning rate becomes zero, as dictated by e.g. Robbins-Monro conditions. However, we find that a learning rate of exactly zero \textit{throughout} training can lead to the fastest convergence in the tabular setting (Fig.~\ref{fig:conv-speed}) when our bounds are applied. 
% We note that this is generally \textit{not} the case for un-regularized RL, as discussed briefly in Appendix~\ref{app:exact}. \JA{I've removed it -- maybe delegate to appendix?}

% In principle, given some assumptions on the structure of the reward function or dynamics, it is possible to tighten these bounds. As an example, we provide a tighter lower bound when the MDP always has an ``identity'' action allowing the agent to return to the same state:
% \TightOnTheFlyBound{}

% In the Appendix, we show that the lower bound of Eq. \ref{eq:tighter_double_sided} is indeed tighter than Eq. \ref{eq:gen_double_sided_boundsA} at all state-actions except the minimizer $(s^*,a^*)=\textrm{arginf}\ \Delta(s,a)$.

As an alternative, in practice, one can replace the $\inf$ and $\sup$ in the previous results by a $\min$ and $\max$, respectively, over some finite dataset (e.g. the current batch of replay data). Although not exact, this substitution becomes increasingly accurate for large datasets (batch sizes), as formalized by our Theorem~\ref{thm:final-prop}. We employ this substitution in the function approximator experiments shown in Section~\ref{sec:experiments}.

After calculating (or estimating) the lower and upper bounds in Theorem~\ref{thm:gen_double_sided_bounds}, we propose to clip the $Q$-function with these bounds at each training step. We conclude this section by showing that the Bellman operator with clipping converges to the optimal $Q$-function:
\begin{tcolorbox}[colback=blue!5!white,colframe=blue!75!black]
\ClippedBOperatorThm{}
\end{tcolorbox}

% \JA{Since the clipped operator converges to the same $Q$-values, our method is unbiased with respect to the value function. We moreover show that this constraint on $Q$ values allows for lower variance in $Q$ and returns during training.} 
This result shows that updates with and without clipping are guaranteed to converge to the same fixed point, $Q^*(s,a)$ (proof in Appendix~\ref{app:exact}). We experimentally demonstrate this statement in Figure~\ref{fig:sql-comparison}. %\edited{Since clipping can only bring the current iterate closer to the optimal $Q$-function, clipped $Q$-learning necessarily has a faster convergence rate, though calculating it (counting the number and magnitude of clips) is generally intractable.}

\subsection{Extension to Continuous Spaces}

The bounds presented in the previous section, though exact, are often intractable due to the required global extremization over continuous state-action spaces. %One cannot access the global extrema of $\Delta$ given only finitely many samples in state-action space. %Thus, we provide the following results, allowing for the extension of our bounds to continuous spaces. 
We therefore loosen the previous bounds by relaxing the required extremization with a simpler optimization over a given batch of replay data.
To this end, we apply the results on Pure Random Search from \citep{malherbe2017global}, bounding the error of estimated extrema of Lipschitz-continuous functions in bounded continuous spaces. We next give a brief discussion on the required steps in the proof, and a full discussion and derivation of Theorem 2 are given in Appendix~\ref{app:cont-bounds}.
% }
%As an example, in the case that one uses the simple upper bound, $Q(s,a) \leq \frac{1}{1-\gamma} \sup r(s,a) $, over a finite-sized batch of replay experience $\{s_i,a_i,r_i, s_{i+1}\}_{i=1}^{n}$, one can bound the (intractable) $\sup$: $\sup r(s,a) \leq \max_i r_i + \mathcal{O}\left(n^{-1/dim(\s) dim(\A)}\right)$. 
We will assume that the extrema of the MDP's reward function $r(s,a)$ can be estimated with high accuracy (since in principle, the entire replay buffer can be used). Instead, we will focus on the larger errors in $\Delta$, which change over the course of training (since we use the most recently learned $Q$-values to generate bounds) and hence has a much smaller dataset available, e.g. the sampled replay batch, for estimation.

Two issues must be addressed before we can apply the concentration results from \citep{malherbe2017global} to Theorem~\ref{thm:gen_double_sided_bounds}: (1) Their concentration analysis requires a Lipschitz constant for the function in question ($\Delta$) which is not readily available; and (2) the exact value of $\Delta$ cannot be given in general, since the dependence on $V(s)$ requires computing an expectation value over states and actions. We surmount these two issues by first providing a calculation for the Lipschitz constant of $\Delta$, based on the Lipschitz constant for a general soft $Q$-function. 
The derivation and discussion of this Lipschitz constant is given in Appendix~\ref{app:lipschitz} due to space constraints. 
% \begin{equation}
%     L_Q=\frac{L_r + \gamma L_p (\beta \sigma_{\min})^{-1}}{1-\gamma L_p (1+L_\mathcal{N})}.
%     \label{eq:L_Q_entreg}
% \end{equation}
% When using a previous solution $Q(s,a)$ to generate double-sided bounds via Theorem \ref{thm:gen_double_sided_bounds}; assuming $Q$ is $L_Q$-Lipschitz and the task of interest is $L_r$-Lipschitz, then one can obtain similar bounds (see Appendix). \JA{"the above result holds"}
Secondly, in the case of stochastic transitions with continuous state-action spaces, we cannot calculate the state-value function term directly. Instead, we apply a concentration inequality to this expectation term, allowing us to bound its error with high probability. A similar error propagation must be considered from the sampling the prior policy in the continuous action setting (to approximate the action expectation in Equation~\ref{eq:soft_bellman}). Hence, with sufficient smoothness and sampling hypotheses, we extend Theorem~\ref{thm:gen_double_sided_bounds} to the sample-based case relevant for the FA setting:

\definecolor{theoremboxcolor}{HTML}{228B22}
% \begin{tcolorbox}[colback=blue!5!white,colframe=blue!75!black]
% \FullPACInformal{}
% \end{tcolorbox}
\begin{tcolorbox}[colback=blue!5!white,colframe=blue!75!black]
\FullPACInformalSimple{}
\end{tcolorbox}

This result shows that our bounds remain valid in the continuous state-action setting with stochastic dynamics, given sufficiently many samples (large batch sizes).
We provide results for other scenarios (discrete or continuous states, deterministic or stochastic transition dynamics), as well as the formal result and relevant definitions in Appendix~\ref{app:cont-bounds}.

\section{Experimental Validation}\label{sec:experiments}
\begin{figure}
    \centering
    \begin{subfigure}[b]{1\textwidth}
        \includegraphics[width=\textwidth]{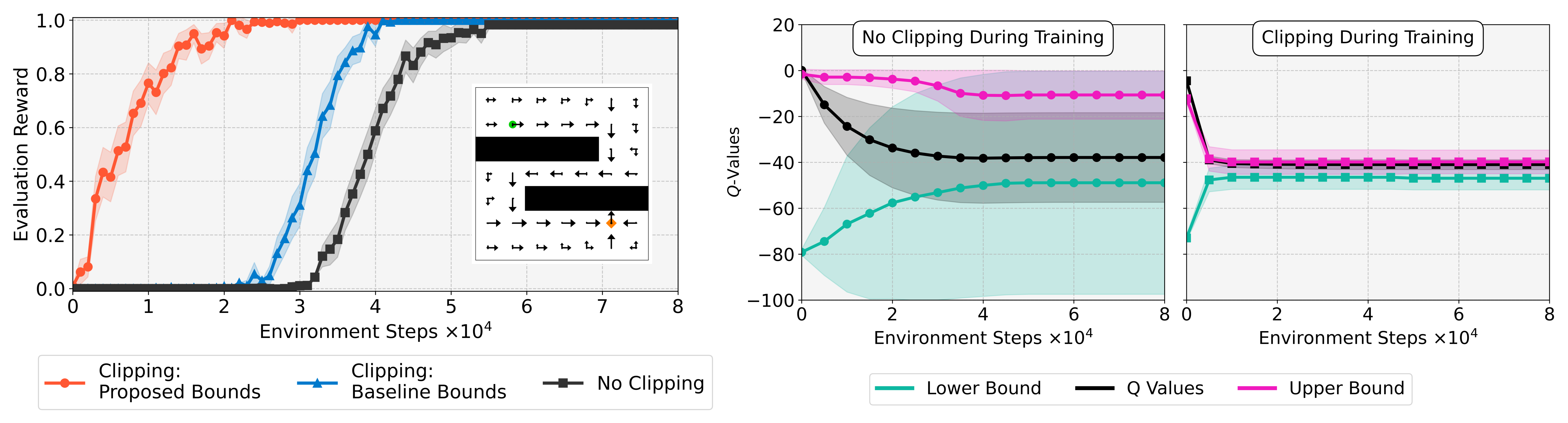}
        % \caption{}
        \label{fig:maze}
    \end{subfigure}
    % \hfill
    % \begin{subfigure}[b]{0.95\textwidth}
    %     \includegraphics[width=\textwidth]{figures/Q_values.png}
    %     \label{fig:maze_learning}
    % \end{subfigure}
    \caption{Here we show specific results on a representative environment, and further examples are given in the Appendix. At each step, the agent receives a small penalty if it has not reached the goal (orange diamond). The discount factor $\gamma=0.98$ and inverse temperature parameter $\beta=5$ are fixed throughout these experiments. From left to right: (1) The optimal policy is shown at the inset. The greedy policy is evaluated during training for the various methods presented. ``Baseline Bounds'' refers to clipping during training with $\left[\frac{r_\textrm{min}}{1-\gamma}, \frac{r_\textrm{max}}{1-\gamma}\right]$. (2,3) The mean and range of $Q$-values and the proposed bounds (Equation~\ref{eq:gen_double_sided_bound}). Clipping during training constrains the $Q$-values to a tight range much faster than without clipping. Each method is averaged over $30$ random initializations.}
    \label{fig:sql-comparison}
\end{figure}

In our experiments, we study the utility of clipping based on our theoretical results.
For simplicity, we first highlight the results in discrete environments with tabular soft $Q$-learning. Without any external estimates for the $Q$-function, we use the estimate given by the previous step's $Q$-values. Note that this method of obtaining an estimate from the previous training step is the most general case, applicable to any value-based algorithm. If available, further information based on other solutions or task structure can additionally be used. Secondly, we study the extension of our theory to the case of continuous states, using the same method for deriving bounds, where we compare clipping methods on the classic control benchmark.

We observe that the use of bounds for clipping the $Q$-function during training leads to a different training dynamics than the standard TD update. Intuitively, clipping restricts the $Q$-function \textit{away} from invalid regions while the Bellman updates pull the $Q$-function \textit{toward} the correct values. The state-action dependency of our bounds also seems to be a key feature, based on comparison to using the looser but constant ``Baseline'' bounds: $Q^*(s,a) \in \left[\frac{R_{\mathrm{min}}}{1-\gamma}, \frac{R_\textrm{max}}{1-\gamma}\right]$ (cf. Figures \ref{fig:sql-comparison} and \ref{fig:conv-speed}).

% \begin{wrapfigure}{R}{0.35\textwidth}
%     \centering
%     \label{fig:conv-speed}
%     \includegraphics[width=\linewidth]{lr_sensitivityavg_reward.png}
%     \caption{Speed of learning (measured as area under evaluation reward curve) with $Q$-table clipping during TD updates. Each point is the result of averaging over 30 random $7\times 7$ mazes with stochastic dynamics.}
% \end{wrapfigure}

% \begin{figure}[h]
%     \centering
%     \includegraphics[width=0.4\linewidth]{lr_sensitivityavg_reward.png}
%     \caption{Speed of learning (measured as area under evaluation reward curve) with $Q$-table clipping during TD updates. Each point is the result of averaging over 30 random $7\times 7$ mazes with stochastic dynamics.}
%     \label{fig:conv-speed}
% \end{figure}
\subsection{Tabular Experiments}
\begin{wrapfigure}{R}{0.4\textwidth}
    \centering
    \includegraphics[width=\linewidth]{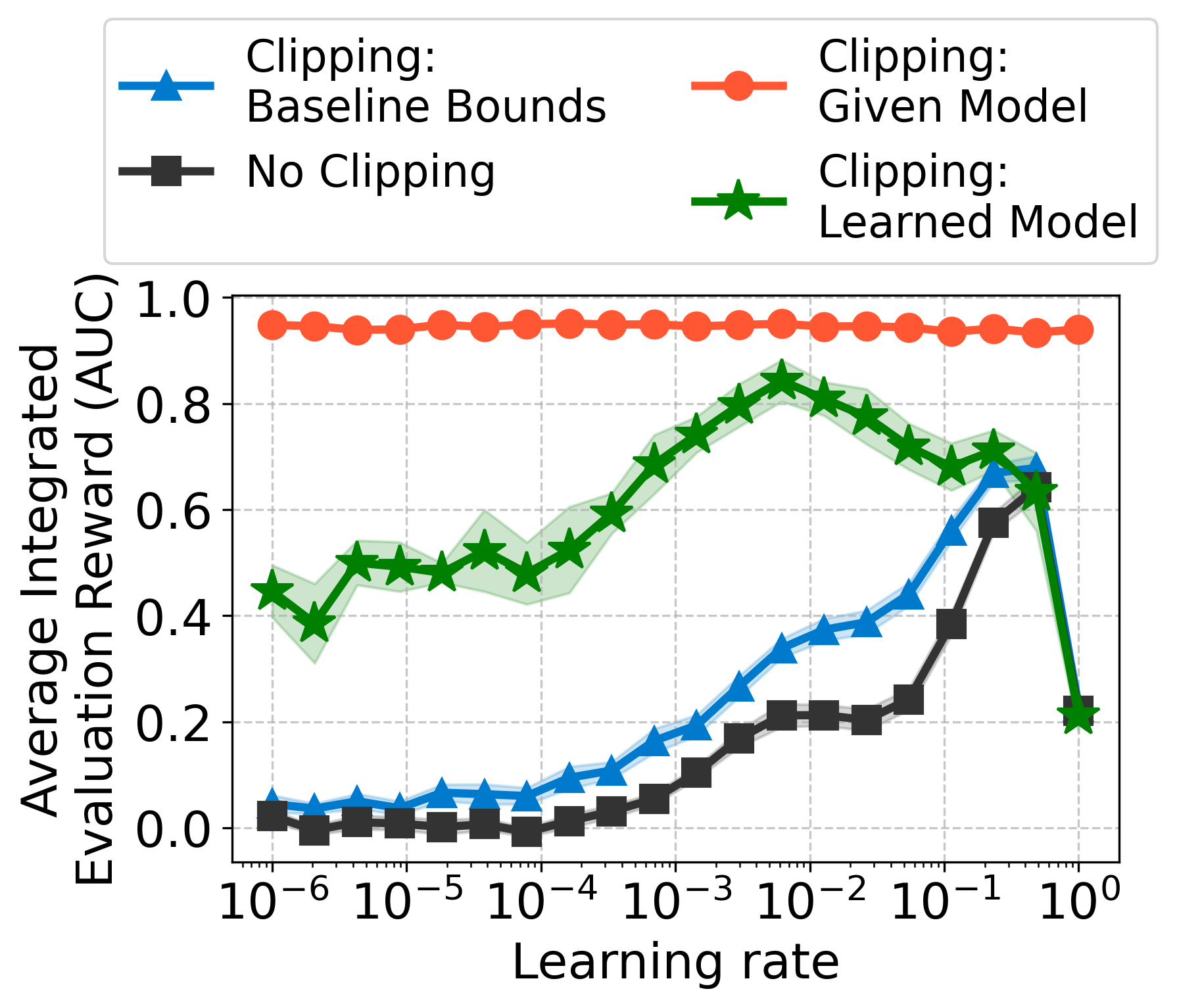}
    \caption{Speed of learning (measured as area under evaluation reward curve) with $Q$-value clipping during TD updates. Each point is the result of averaging over 30 randomly generated $7\times 7$ mazes with stochastic transitions. Further details of the experiment are given in Appendix~\ref{app:code}.}
    \label{fig:conv-speed}
\end{wrapfigure}In the tabular case, since we have access to the $Q$-table, we can simply clip the updated $Q$-table according to the derived bounds. When the model (reward and transition table) is given, we maintain the lower and upper bounds throughout training, tightening them whenever a better bound becomes available. In Figure \ref{fig:sql-comparison} we show the results of training in a simple maze environment. In the main plots of Figure \ref{fig:sql-comparison}, we depict a comparison of the evaluation rewards and the mean $Q$-value over all $(s,a)$ pairs.
%%% The pat below doesn't make sense, so I commented it out. Please rephrase if important.
%after introduced a more model-free approach, closer to the case of FAs considered later.} 
In experiments across different sized environments, and with various levels of stochasticity, we universally find the increase in convergence speed shown. 

To verify the robustness of clipping benefits, we sweep over the learning rate and maze topology, and measure the speed of convergence: plotted in Figure~\ref{fig:conv-speed}. The metric used is a normalized area under the evaluation reward curve. Since randomly generated mazes are used, we normalize against the performance of a uniform policy and the greedy optimal policy (given by the exact solution). Thus an algorithm which quickly obtains (and maintains) the optimal reward will have a larger success metric (more details in Appendix~\ref{app:code}). 
In this experiment, we use stochastic transition dynamics. %The entire range of learning rates considered outperforms the ``No Clipping'' baseline by a significant margin. 
We begin by giving the agent the model (in this case, simply a table of rewards and stochastic transition probabilities), which is required for an exact calculation of the bounds in Theorem~\ref{thm:gen_double_sided_bounds}. This setting of a ``Given Model'' is shown with red circles in Figure~\ref{fig:conv-speed}, which outperforms the other methods, for all learning rates. Then, we consider the case of a learned model, shown with green stars in Figure~\ref{fig:conv-speed}. This case represents a stepping stone from exact tabular updates (red line) to the function approximator case since there is noise in the calculation of $\Delta(s,a)$ based on sampling errors introduced by the learned model.
%This suggests that only clipping the $Q$-table with our bounds is able to provide a more efficient learning signal without explicit use of Bellman target updates.\footnote{Although it is worth noting the Bellman loss is propagated implicitly through $\Delta$.} 

We have found that there are initializations for which our bounds are not violated and thus the $Q$-function is not (initially) clipped by our bounds.
%the $Q$-function is not initially clipped by our bounds for all initializations.
For instance, we find that for $Q$-values initialized far from zero, the bounds are loose and not violated, hence small learning rates will yield nearly static training dynamics which do not converge. This observation, coupled with the 
finding that updating via clipping (when the bounds \textit{are} violated)
%finding that lower learning rates with clipping 
consistently performs better (red line with circle markers in Fig. \ref{fig:conv-speed}) leads us to the following: To take advantage of the boosting given by clipping in the low learning rate regime, we propose to use the standard temporal difference (TD) update only if the $Q$-function has not changed between two iterations (i.e. no clipping occurs), and otherwise clip the $Q$-values appropriately \textit{without any TD update}. This should be distinguished from a simpler ``always clip'' approach, shown in Algorithm~\ref{alg:always-clip}. This change ensures convergence of a tabular SQL algorithm while maximizing the utility of clipping. Pseudocode for the algorithm is  given in Algorithm~\ref{alg:clip} in Appendix~\ref{app:code}. The performance difference between Algorithms~\ref{alg:always-clip} and~\ref{alg:clip} is shown in Figure~\ref{fig:alg_comparison} in Appendix~\ref{app:expts}. Two versions (one with the model given and one with a learned model) of our proposed algorithm are shown in Figure~\ref{fig:conv-speed}.  
% Interestingly, we find that (averaged across 30 randomly generated $7\times 7$ mazes with stochastic dynamics) using our Algorithm~\ref{alg:clip} yields a considerable performance increase. Moreover, we find that (with proper initialization), the Bellman update (highlighted in blue) is never triggered, and clipping alone is able to push the $Q$-values to convergence, faster than with any learning rate and Bellman updates alone, shown in Fig.~\ref{fig:conv-speed}

\subsection{Function Approximator Experiments}
\begin{figure}[t]
    \centering
    \includegraphics[width=1\textwidth]{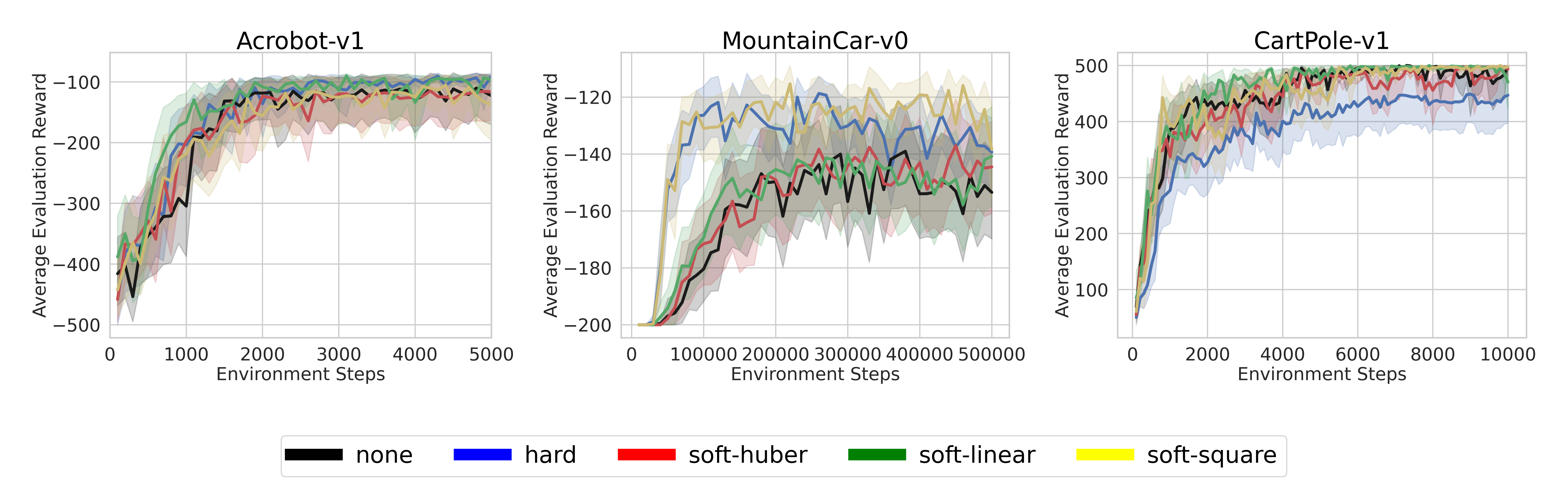}
    \caption{We test the proposed clipping methods (labeled None, Hard, and Soft; described below) across the classic control suite. We fine-tuned each environment's hyperparameters (details in Appendix~\ref{app:code}). The average evaluation reward plotted is the reward achieved by following the stochastic optimal policy, averaged over 5 episodes. Each method in a given environment is averaged over 30 random initializations, with the 95\% bootstrapped confidence interval shaded. To ensure the performance stems from our bounds alone, we have not included the simpler $R_\text{min,max}/(1-\gamma)$ bounds which are likely to improve the performance further.}
    \label{fig:benchmark}
\end{figure}

To test our bounds in the deep RL setting, we turn to environments with continuous state spaces, adopting a DQN-style implementation of discrete-action soft $Q$-learning with an entropy-regularized TD update, using an online and target network.
Although we have derived bounds for this case, we cannot simply clip the entire $Q$-function as we did in the tabular setting. The proposed algorithm (with clipping only if the $Q$-function remains fixed) cannot be directly translated to the FA setting. Thus, we will propose two different methods of clipping suitable for function approximators.

Since the soft $Q$-learning (SQL) algorithm employs a target network we use both the target and online networks to derive bounds on the optimal $Q$-values (cf. Appendix~\ref{app:expts} for the implementation details and hyperparameters used for training). Since the bounds must hold when either the target or online net is used as an estimate, we can always take the tighter bound $(s,a)$-wise between the two. In general, given many sources of $Q$-function estimates (such as in ensemble methods), one can use them collectively to obtain the tightest bound possible.

The derived bounds can be implemented using different approaches for clipping of the value function during training. We highlight the different methods used below, inspired by the methods used by \cite{he2017learning, constrainedGPI, Adamczyk_UAI}:
\newline
(0) \textbf{No Clipping:} The standard training scheme for SQL is implemented. %We nevertheless monitor the bound violations and fraction of updates that otherwise \textit{would be} clipped during training.
\newline
(1) \textbf{Hard Clipping:} At each backup calculation we enforce the following bounds on the new $Q$-value:
% \begin{equation}
%      \hat{Q}_{\textrm{clip}}(s,a) 
% \end{equation}
% where 
\begin{equation}
    Q(s,a) \xleftarrow[]{}\hat{Q}_{\textrm{clip}}(s,a) \doteq \min\left\{ \max\left\{ r(s,a) + \frac{\gamma}{\beta} \log \E_{a' \sim{} \pi_0} \exp \beta Q(s',a'),\  \text{L}(s,a) \right\}, \text{U}(s,a) \right\}
\label{eq:qclip}
\end{equation}
and $\text{L}$ and $\text{U}$ denote the lower and upper bounds derived in Theorem \ref{thm:gen_double_sided_bounds}. In the tabular setting, $\text{L}$~and~$\text{U}$~can be calculated exactly. However, for function approximator experiments with sampling, we replace the $\inf$ and $\sup$ with a $\min$ and $\max$ over the current batch as justified in Section~\ref{sec:Theory}.
\newline
(2) \textbf{Soft Clipping:} An additional term, the ``clipping loss'', is added to the function approximator's loss function. The clipping loss is defined as 
\begin{equation}
    % \mathcal{L}_{\textrm{clip}}=|Q(s,a) - \min\{ \max\{ Q(s,a),\  \text{L}(s,a) \}, \text{U}(s,a) \}|,
    \mathcal{L}_{\textrm{clip}}= \frac{1}{|\mathcal{B}|}\sum \av{Q(s,a) - \hat{Q}_{\textrm{clip}}(s,a)},
    \label{eq:clip_loss}
\end{equation}
with a summation over the current batch, where $|\mathcal{B}|$ is the batch size. 
This gives a total loss of $\mathcal{L}=\mathcal{L}_{\text{Bellman}} + \eta \mathcal{L}_{\text{clip}}$. The hyperparameter $\eta$ weights the relative importance of the bound violations against the Bellman error, whose value we fix to unity for simplicity. On an environment-wise basis, we fine-tune the hyperparameters on the baseline method (values in Appendix~\ref{app:hparams}) and use those same values for all clip methods. %Alternatively, one can view this additional loss as being equivalent to a reward bonus for states with high bound violation. This is analogous to the UCB-style bonus applied in \cite{lee2021sunrise}. 
Figure \ref{fig:benchmark} indicates that clipping can lead to improvements in the speed of training; however, additional modifications are needed to further validate the benefits of clipping in the FA setting. In Figure~\ref{fig:benchmark}, we observe that replacing the $\ell_1$-loss shown in Equation~\ref{eq:clip_loss} with the Huber or $\ell_2$ loss can yield better performance depending on the environment. However, this effect can likely be mitigated by fine-tuning the weight {parameter,~$\eta$}.
\iffalse
\newline
(3) \textbf{Smoothed Clipping:} The updated $Q$-values are set as an average between those given by Hard Clipping and No Clipping, with a relative weight factor inversely related to the bound violations.
\begin{align*}
    Q(s,a) \xrightarrow[]{} (1-\tau)\left(r(s,a) + \gamma V(s')\right) + 
    \tau \hat{Q}_{\textrm{clip}}(s,a)
\end{align*}
where 
\begin{equation}
    \tau = \frac{\mathcal{L}_{\text{clip}}}{1+\mathcal{L}_{\text{clip}}}
\end{equation}

We note that when the bound violations are zero, the standard update rule is recovered. This value for $\tau$ is chosen to set the relative weight of the two terms to match the magnitude of bound violations: $\tau/(1-\tau)=\mathcal{L}_{\text{clip}}$. Therefore, the clipped values will be preferred over the standard update rule, in direct proportion to the bound violations.
\fi

\section{Discussion}

In this work, we have given a theoretical foundation for deriving double-sided bounds in reinforcement learning, showcasing their experimental validity. Our investigation in tabular domains has demonstrated that application of these bounds significantly boosts training speed. Coupling our bounds with proof techniques in e.g. \citep{tang2020self} may allow for a proof of a faster convergence rate.
Beyond the theoretical contributions, our work calls for exploration in several new research directions. While our derived bounds hold in general, there is potential for further refinement given specific classes of value function estimates and transition dynamics or reward function structures, as discussed in Section~\ref{sec:prior}. There is also the potential in transfer learning to leverage bound violation minimization at a state-action level to construct refined initializations from a diverse set of policies.

Integrating our results with other state-of-the-art methods in value-based learning seems a promising direction for future study. Several specific examples include: exploiting ensembles and extending to continuous actor-critic methods, adopting a dynamic schedule for the soft clipping weight parameter, akin to that in \citep{Haarnoja_SAC}, and interfacing with model-based approaches such as DYNA~\citep{dyna}, where our tabular results suggest that a more significant performance boost may be achieved. We believe that integrating these methods is an important step to ensuring utility in more complex environments. Finally, we note an intriguing suggestion arising from our experiments, which can be loosely summarized as ``clipping is all you need''. Further translating the benefits of clipping from tabular to deep RL presents an exciting opportunity for future research.

\subsubsection*{Acknowledgments}
\label{sec:ack}
JA would like to acknowledge the use of the supercomputing facilities managed by the Research Computing Department at UMass Boston. The work of JA was supported in part by the College of Science and Mathematics Dean's Doctoral Research Fellowship through support from Oracle, project ID R20000000025727; and in part by the Alliance Innovation Lab -- Silicon Valley. RVK and JA would like to acknowledge funding support from the NSF through Award No. DMS-1854350. ST and VM were supported in part by the NSF Award No. 2246221, the Pazy Foundation grant No. 195-2020, and the Alliance Innovation Lab -- Silicon Valley. 

%%%%%%%%%%%%%%%%%%%%%%%%%%%%%%%%%%%%%%%%%%%%%%%%%%%%%%%%%%%%%%%%
%% Bibliography
%%%%%%%%%%%%%%%%%%%%%%%%%%%%%%%%%%%%%%%%%%%%%%%%%%%%%%%%%%%%%%%%
\bibliography{main}
\bibliographystyle{rlc}

%%%%%%%%%%%%%%%%%%%%%%%%%%%%%%%%%%%%%%%%%%%%%%%%%%%%%%%%%%%%%%%%
%% Appendices
%%%%%%%%%%%%%%%%%%%%%%%%%%%%%%%%%%%%%%%%%%%%%%%%%%%%%%%%%%%%%%%%
\clearpage
\appendix

% \section{Technical Appendix}
% In this technical appendix, we provide further discussion on experimental details and give proofs for all the results shown in the main text. 
% \onecolumn
% \iftoggle{appendix}{
% \edited{add more intro here}
\section{Experiments\label{app:expts}} 
% \subsection{Function Approximator Experiments}
In tabular soft $Q$-learning, we calculate the Bellman residual, mixing it into the current estimate of $Q(s,a)$ at every step taken by the agent. At each update step, we calculate the bounds given by Theorem \ref{thm:gen_double_sided_bounds}, which are exact in the case that the model is given (red circles in Figure~\ref{fig:conv-speed}). Since these bounds are exact, we can repeatedly take the tightest possible bounds at every step, leading to the consistent fast convergence. When the model is learned through sampling (updating the deterministic reward table and using count-based estimates for the stochastic transition dynamics), the bounds are inexact, so we only use the current step's estimate without iteratively tightening them, which we found to lead to collapse to incorrect values. The clipping performed follows Equation~\ref{eq:qclip} in the main text. %\JA{Interestingly, we see that as the upper bound becomes tight, the $Q$-values are constantly saturated by this value. The departure of the No Clipping and Hard Clipping $Q$-values is also evident in the reduction of error ($\ell_\infty$ distance) between consecutive iterations.}

\begin{figure}
    \centering
    \includegraphics[width=1\textwidth]{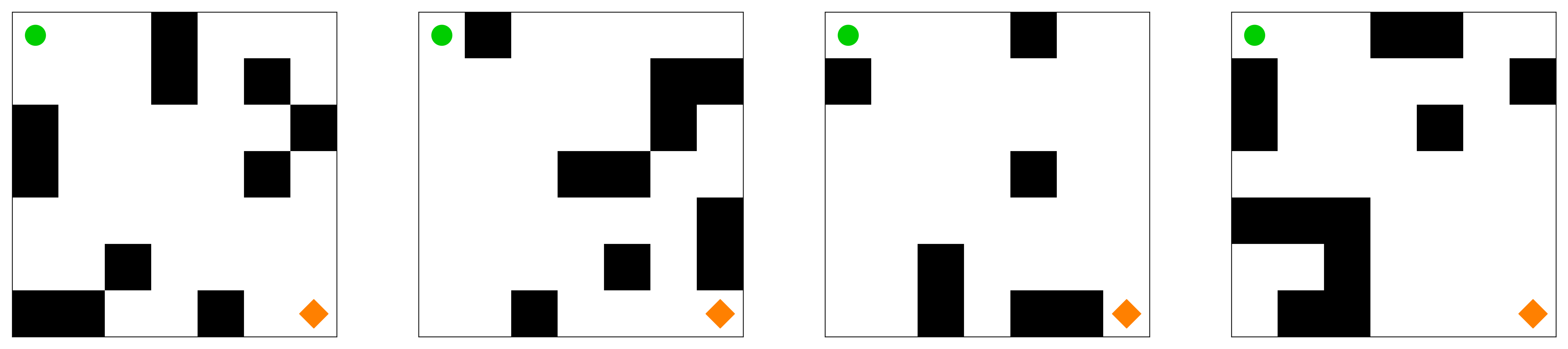}
    \caption{Examples of the random maps generated for the tabular experiments.}
    \label{fig:ex-mazes}
\end{figure}

For the tabular experiments, we first generated $30$ random mazes for each method to solve $10$ times. In each generated $7\times7$ maze, walls are randomly generated at a site with probability $20\%$, and a goal is randomly placed at a site without a wall. We show four examples of such mazes in Figure~\ref{fig:ex-mazes}. Depth-first search is used to ensure the generated maze has a valid solution (i.e., the rewarding state can be visited by the agent). Each step costs the agent $-1$, and the goal state incurs a cost of $-0.25$. The environment is stochastic, such that the probability of moving in the ``intended'' direction is 75\%, and the probability of moving perpendicular to the intended direction is 12.5\%. The agent then transitions to the grid state one unit in that direction, as common in MiniGrid or FrozenLake environments. Although we have sparse rewards for the simplicity of environment generation, we find our results to hold across various dense reward settings, with varying levels of stochasticity. 

As mentioned in the main text, the ability for bounds to be applied (and training efficiency overall) is sensitive to the scale of the initialized $Q$-function. We found that a random uniform initialization of the $Q$-values in the range $(-1,1)$ performs best for the baseline, and thus we maintain this initialization across all experiments.

Since each maze potentially has a different evaluation reward scale, we normalize the evaluation score so they may be averaged across mazes, akin to e.g. \cite{mnih2015human}:
\begin{equation}
    \textrm{Normalized Evaluation Reward} = \frac{\langle R\rangle_\text{optimal}-\langle R\rangle_\text{agent}}{\langle R\rangle_\text{agent} - \langle R\rangle_{\text{uniform}}},
\end{equation}

where $\langle R\rangle_\text{optimal},\ \langle R\rangle_\text{agent},\ \langle R\rangle_\text{uniform}$ denote the average reward (over 3 episodes) obtained by an agent executing an optimal, training, or uniform random policy, respectively. Finally, we integrate the area under this ``Normalized Evaluation Reward'' vs. Environment Steps curve, to obtain a metric for the speed of convergence, plotted in Figure~\ref{fig:conv-speed}: ``Average Integrated Evaluation Reward~(AUC)''.

To explore the utility of clipping in function approximator (FA) systems, we use a soft Q-learning (SQL) algorithm \cite{haarnoja2017reinforcement}, while applying and monitoring clipping given by the bounds in Theorem \ref{thm:gen_double_sided_bounds}. In particular, we continuously bootstrap by using the previous estimate of the $Q$-function to generate the bounds, and we clip the target network's output value accordingly. More specifically, we extract bounds from both the target network and $Q$-network at each step, and take the tighter of the two bounds. For continuous spaces, we use the estimate $\sup r(s,a) \approx \max_{i \in \mathcal{D}} r(s,a)$, where the $\max$ is taken over the current batch (and similarly for $\inf r(s,a)$). We consider the two clipping methods described in the Experiments section of the main text. We have also experimented with different loss functions, as the optimal choice seems to be environment-dependent. %As mentioned in the main text, a single clip-loss function could be chosen, at the expense of tuning the additional hyperparameter $\eta(=1)$.

\subsection{Implementation Details\label{app:code}}
For many environments of interest, the transition dynamics are reducible (they have absorbing states returning a termination signal (``done'' in Gym~\cite{openAI},  ``terminated'' in the newer Gymnasium package~\cite{gymnasium}). A common method to assign a $Q$-value to such states is given by (see e.g. \cite{mnih2015human}):
\begin{equation}
    Q(s,a) = \begin{cases} 
r(s, a) & \text{if terminated} \\
r(s, a) + \gamma V(s') & \text{else} 
\end{cases}
\label{eq:convention}
\end{equation}
This value assignment means that states near absorbing states will have values $\sim{} \mathcal{O}(1)$ rather than the $\mathcal{O}(\frac{1}{1-\gamma})$ given by the bounds presented. In passing, we note that one way to circumvent this would be to alter the convention by assigning a value of $r(s,a)/(1-\gamma)$ at termination, since for irreducible dynamics, this would correspond to accumulating the terminal state's reward \textit{ad infinitum}.

To conform to the convention shown above, we modify our bounds to allow for the termination signal to properly affect the bounds. Focusing on deterministic dynamics for simplicity, the bounds are modified from:
\begin{align*}
    Q^*(s,a) &\leq r(s,a)+\gamma \biggr(\hat{V}(s') + \frac{\max_{(s,a) \in \mathcal{B}} \hat{\Delta}(s,a) }{1-\gamma} \biggr) \\
    Q^*(s,a) &\geq  r(s,a)+\gamma \biggr(\hat{V}(s') + \frac{\min_{(s,a) \in \mathcal{B}} \hat{\Delta}(s,a) }{1-\gamma} \biggr) 
\end{align*}
to the following:
\begin{align*}
    Q^*(s,a) &\leq r(s,a)+\gamma \biggr(\hat{V}(s') + \frac{\max_{(s,a) \in \mathcal{B}} \hat{\Delta}(s,a) }{1-\gamma} \biggr)[1-\mathrm{done(s')}] \\
    Q^*(s,a) &\geq  r(s,a)+\gamma \biggr(\hat{V}(s') + \frac{\min_{(s,a) \in \mathcal{B}} \hat{\Delta}(s,a) }{1-\gamma} \biggr)[1-\mathrm{done(s')}]
\end{align*}
which can be justified by referring to the value definition used (Equation~\ref{eq:convention}).

% We can further tighten the bounds in the case of absorbing dynamics by noting that the $\hat{\Delta}/(1-\gamma)$ contribution for the next state should only be considered over states which are not terminal:
% \begin{equation}
%     \mathcal{B}  \to \mathcal{B}\ \backslash\ \{(s,a) : \mathrm{done}(s')=\mathrm{True} \}
% \end{equation}

% where $\mathrm{done}(s')$ denotes the next-state being an absorbing state.
\clearpage
{
Next we provide the algorithm used for clipping in our experiments. We highlight in {\color{blue}blue} the changes from a standard soft $Q$-learning approach without clipping, e.g. a discrete-action analogue of \cite{haarnoja2017reinforcement}.

\begin{algorithm}
\caption{Soft $Q$-learning with Constant Clipping (Tabular)}
\begin{algorithmic}[1]
    \State Initialize $Q$-values: $Q(s, a) \sim \mathrm{Unif}(-1,1)$, max sample budget.
    \State Initialize $\mathrm{L}(s,a) = -\infty$, $\mathrm{U}(s,a)=+\infty$.
    \State Set learning rate $\alpha$, discount factor $\gamma$, and inverse temperature $\beta$.
    \While{total environment steps $<$ max sample budget}
        \State Reset environment
        \While{not end of episode}
            \State Choose action $a \sim \pi(\cdot|s) \propto \exp{\beta Q(s, a)}$
            \State Take action $a$: observe reward $r$, next state $s'$, and termination signal
            \State Compute state value function:
            $V(s')=\beta^{-1} \log \E_{a'\sim{} \pi_0} \exp \beta Q(s',a')$
            \State Compute the TD error: $\delta = r + \gamma \cdot (1 - \mathrm{terminated}) \cdot V(s') - Q(s, a)$
                \State Update $Q$-table: $Q'(s, a) = Q(s, a) + \alpha \delta$
            {\color{blue}
            \State Calculate new bounds $\{\mathrm{L}'(s,a),\ \mathrm{U}'(s,a)\}$ using $Q'$ in Theorem~\ref{thm:gen_double_sided_bounds}.
            \State Tighten lower bounds: $\mathrm{L}'(s,a) = \max \left\{\mathrm{L}'(s,a),\ \mathrm{L}(s,a)\right\}$
            \State Tighten upper bounds: $\mathrm{U}'(s,a) = \min \left\{\mathrm{U}'(s,a),\ \mathrm{U}(s,a)\right\}$
            \State Clip the $Q$-values: $Q'(s,a)=\mathrm{clamp}\left(Q'(s,a), \min = \mathrm{L}'(s,a), \max = \mathrm{U}'(s,a)\right)$
            }
            \State Update state: $s \gets s'$
            \State Update $Q$: $Q \gets Q'$
        \EndWhile
    \EndWhile
\end{algorithmic}
\label{alg:always-clip}
\end{algorithm}
}

{
% \begin{figure}[h]
\begin{algorithm}
\caption{Soft Q-learning with Conditional TD-updates (Tabular)}
\begin{algorithmic}[1]
    \State Initialize $Q$-values: $Q(s, a) \sim \mathrm{Unif}(-1,1)$, max sample budget.
    \State Initialize $\mathrm{L}(s,a) = -\infty$, $\mathrm{U}(s,a)=+\infty$.
    \State Set learning rate $\alpha$, discount factor $\gamma$, and inverse temperature $\beta$.
    \While{total environment steps $<$ max sample budget}
        \State Reset environment
        \While{not end of episode}
            \State Choose action $a \sim \pi(\cdot|s) \propto \exp{\beta Q(s, a)}$
            \State Take action $a$: observe reward $r$, next state $s'$, and termination signal
            \State Compute state value function:
            $V(s')=\beta^{-1} \log \E_{a'\sim{} \pi_0} \exp \beta Q(s',a')$
            {\color{blue}\State Calculate new bounds $\{\mathrm{L}'(s,a),\ \mathrm{U}'(s,a)\}$ using $Q'$ in Equation~\ref{eq:gen_double_sided_bound}.
            \State Tighten lower bounds: $\mathrm{L}'(s,a) = \max \left\{\mathrm{L}'(s,a),\ \mathrm{L}(s,a)\right\}$
            \State Tighten upper bounds: $\mathrm{U}'(s,a) = \min \left\{\mathrm{U}'(s,a),\ \mathrm{U}(s,a)\right\}$
            \State Clip the $Q$-values: $Q'(s,a)=\mathrm{clamp}\left(Q(s,a), \min = \mathrm{L}'(s,a), \max = \mathrm{U}'(s,a)\right)$
            {\If{$Q' == Q$ }
                \State // No clipping has been applied, resort to TD-update:
                \State Compute the TD error: $\delta = r + \gamma \cdot (1 - \mathrm{terminated}) \cdot V(s') - Q(s, a)$
                \State Update $Q$-table: $Q'(s, a) \gets Q'(s, a) + \alpha \delta$
            \EndIf
            }
            } 
        \State \hspace{-\parindent} Update state: $s \gets s'$
        \State \hspace{-\parindent} Update $Q$: $Q \gets Q'$
    \EndWhile
\EndWhile
\end{algorithmic}
\label{alg:clip}
\end{algorithm}
% \caption{this algor...}
% \end{figure}

\begin{figure}
    \centering
    \includegraphics[width=0.6\textwidth]{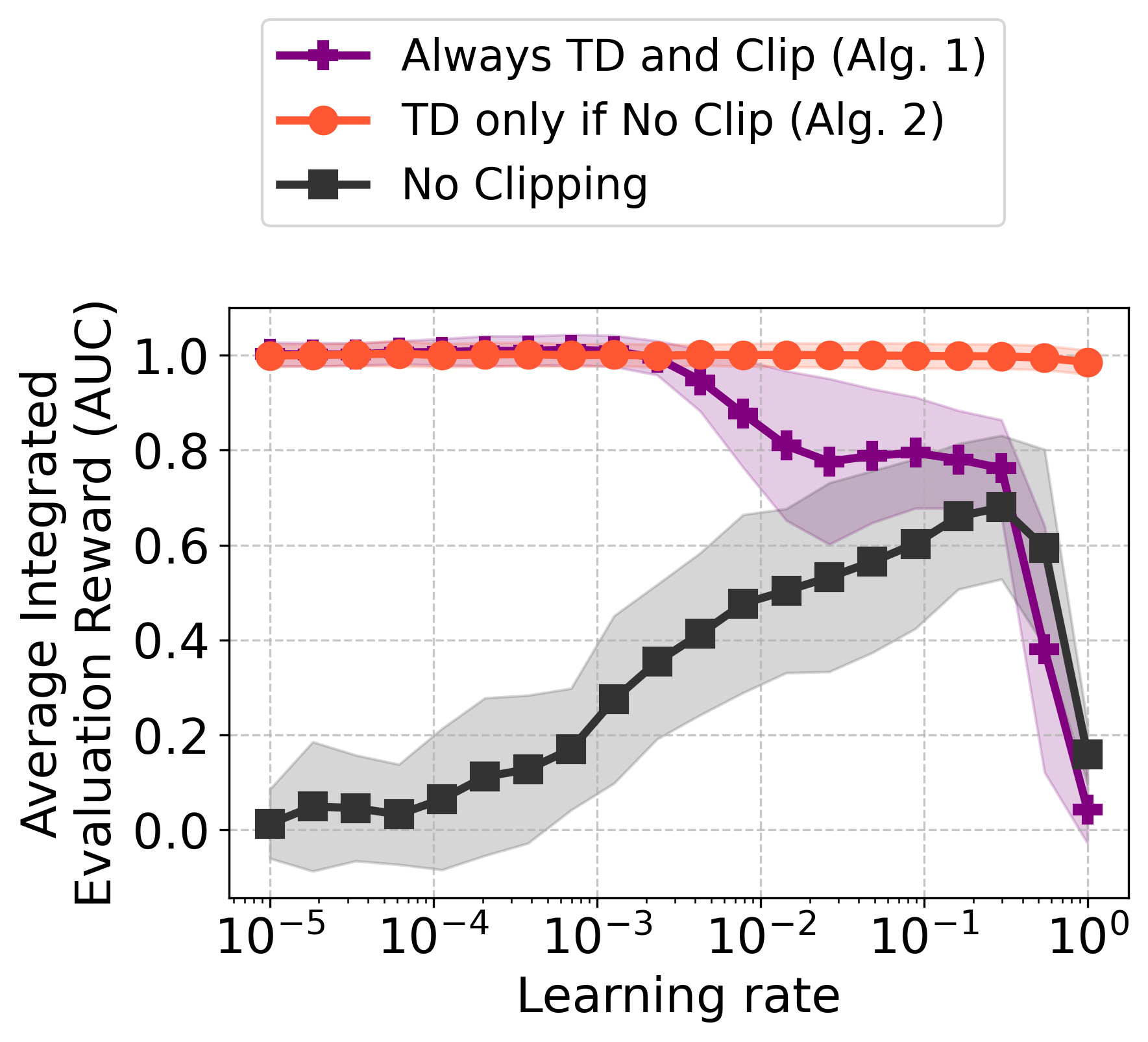}
    \caption{In the $7 \times 7$ mazes, we compare the ``always clip'' (Algorithm~\ref{alg:always-clip}) and ``TD only if no clipping'' (Algorithm~\ref{alg:clip}) algorithms as discussed in Section~\ref{sec:experiments}. The points labeled ``TD only if No Clip'' represent the same algorithm shown in the main text's Figure~\ref{fig:conv-speed}, titled ``Clipping: Given Model''.}
    \label{fig:alg_comparison}
\end{figure}
}

In Figure~\ref{fig:alg_comparison}, we compare Alg.~\ref{alg:always-clip} and Alg.~\ref{alg:clip}. Importantly, we find it is imperative to not \textit{always} update the $Q$-function with TD updates. Rather, we find that by using TD updates only when the $Q$-values are not changed by clipping, the performance significantly improves in the high learning rate regime. %\edited{However, since we do not have full access to the $Q$-function nor successive batches, further investigation is needed to implement Algorithm~\ref{alg:clip} in deep RL.}

\begin{figure}
    \centering
    \includegraphics[width=0.6\textwidth]{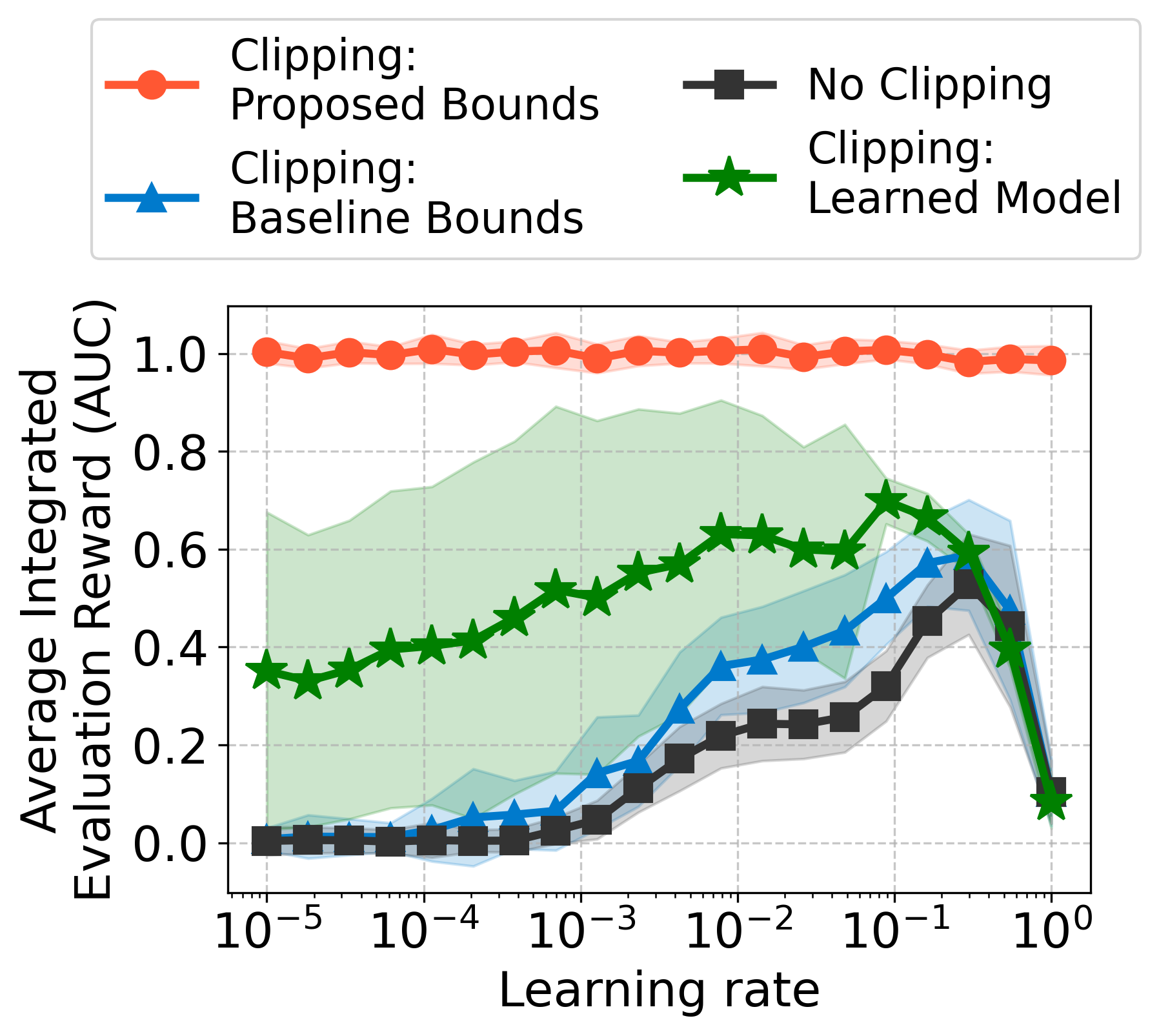}
    \caption{We perform the same experiments as demonstrated in Figure~\ref{fig:conv-speed}, on larger $30 \times 30$ mazes, with the same qualitative results.}
    \label{fig:my_label}
\end{figure}

\clearpage
\subsection{Hyperparameters\label{app:hparams}}
For the FA classic control experiments, we parameterize the $Q$-function by an MLP with a standard fixed depth (two hidden layers) and fine-tuned width. We keep the discount factor $\gamma=0.99$ fixed across tasks, and use a single online and target function. We tune the learning rate, $\beta$, target update frequency, training frequency, number of gradient steps per training step, and batch size. The ranges for each hyperparameter, swept uniformly at random, are given below:

\begin{table}[htbp]
    \centering
    \caption{Hyperparameters and Ranges}
    \label{tab:hyperparameters}
    \begin{tabular}{lll}
        \toprule
        Hyperparameter & Range & Sampling Distribution \\
        \midrule
        Learning Rate & $(10^{-4}, 10^{-1})$ & Log Uniform \\
        Inverse Temperature, $\beta$ & $(10^{-2}, 10^{1})$ & Log Uniform \\
        Target Update Frequency & $\left\{1,10,100,1000\right\}$ & Uniform \\
        Training Frequency & $(1,100)$ & Log Uniform \\
        Gradient Steps per Training Step & $(1,100)$ & Log Uniform \\
        Batch Size & $\left\{2^4, 2^5, 2^6, 2^7, 2^8, 2^9, 2^{10}\right\}$ & Uniform \\
        % Soft Weight & $(0.001, 50)$ \\
        Hidden Dimension & $\left\{2^4, 2^5, 2^6, 2^7, 2^8, 2^9 \right\}$ & Uniform \\
        \bottomrule
    \end{tabular}
    \caption*{We sweep each hyperparameter at random in the ranges shown, and select the best hyperparameter set, sorted by the largest area under the evaluation reward curve (averaged over 3 independent runs). The best hyperparameters for each environment are shown in the next table.}
\end{table}

\begin{table}[htbp]
    \centering
    \caption{fine-tuned Hyperparameters for No Clipping (Baseline) Soft $Q$-Learning}
    \label{tab:fine_tuned_hyperparameters_all}
    \begin{tabular}{lllllllll}
        \toprule
        \midrule
        Environment & CartPole-v1 & Acrobot-v1 & MountainCar-v0  \\
        \midrule
        Learning Rate & $0.016$ & $0.0005$ & $0.007$ \\
        Inverse Temperature, $\beta$ & $0.019$ & $4.5$ & $5.3$ \\
        Target Update Frequency & $1$ & $10$ & $10$  \\
        Training Frequency & $2$ & $2$ & $58$  \\
        Gradient Steps per Training Step & $16$ & $20$ & $5$\\
        Batch Size & $512$ & $64$ & $128$\\
        Learning Starts & 0 & 0 & 10,000 \\
        % Soft Weight & $10$ & $5$ & $2$ & $0.001$ & $20$ \\
        Hidden Dimension & 64 & 64 & 512 \\
        \bottomrule
    \end{tabular}
\end{table}

\clearpage 
\newpage
\section{Proofs of Exact Results\label{app:exact}}
In this section and the next we provide proofs of the theoretical results in the main text.
Each proof is prefaced with a restatement of the theorem for convenience.

We begin with a helpful lemma which bounds the optimal action-value function $Q^*(s,a)$ for any task.
We note that these bounds hold for both un-regularized RL and entropy-regularized RL.
\begin{tcolorbox}[colback=blue!5!white,colframe=blue!75!black]
\begin{customlemma}{A}
    For a task with reward function $r(s,a)$, discount factor $\gamma$, the (soft) optimal action-value function $Q^*(s,a)$ satisfies:
    \begin{align}
        Q^*(s,a) &\geq r(s,a) + \gamma \frac{\inf_{s,a} r(s,a)}{1 - \gamma} \label{eq:lb}\\
        Q^*(s,a) &\leq r(s,a) + \gamma \frac{\sup_{s,a} r(s,a)}{1 - \gamma} \label{eq:ub}
    \end{align}
    \label{lem:double_sided_bound}
\end{customlemma}
\end{tcolorbox}

\begin{proof}
We will prove the upper bound here, with the lower bound's proof following similarly. The proof follows from induction on steps ($n$) of the recursive Bellman backup equation:
    \begin{equation}
        Q^{(n+1)}(s,a) = r(s,a) + \frac{\gamma}{\beta}\E_{s' \sim{} p(\cdot | s,a)} \log \E_{a'\sim{} \pi_0(\cdot|s')} \exp \left( \beta Q^{(n)}(s',a') \right).
    \end{equation}
    We first use induction to prove
    \begin{align*}
        % Q^{(n)}(s,a) &\geq r(s,a) + \gamma \frac{1-\gamma^n}{1 - \gamma}\inf_{s,a} r(s,a)\\
        Q^{(n)}(s,a) &\leq r(s,a) + \gamma \frac{1-\gamma^n}{1 - \gamma}\sup_{s,a} r(s,a).
    \end{align*}
    Then, since $\lim_{n \to \infty}Q^{(n)}(s,a)=Q^*(s,a)$ and $\gamma \in [0,1)$ the desired result (Equation~\ref{eq:ub}) will follow from this limit.

    We set $Q^{(0)}(s,a)=r(s,a)$. The base case ($n=1$) trivially holds:
    \begin{align*}
        Q^{(1)}(s,a) &= r(s,a) + \frac{\gamma}{\beta}\E_{s' \sim{} p(\cdot | s,a)} \log \E_{a'\sim{} \pi_0(\cdot|s')} \exp \left( \beta Q^{(0)}(s',a') \right) \\
        &= r(s,a) + \frac{\gamma}{\beta}\E_{s' \sim{} p(\cdot | s,a)} \log \E_{a'\sim{} \pi_0(\cdot|s')} \exp \left( \beta r(s',a') \right) \\
        &\leq r(s,a) + \frac{\gamma}{\beta}\sup_{s,a} \left( \beta r(s,a) \right) \\
        &= r(s,a) + \gamma \frac{1-\gamma^{1}}{1-\gamma}\sup_{s,a}r(s,a).
    \end{align*}
    We proceed in proving the upper bound based on induction as described above.
    For notational convenience we denote $R \doteq \sup r(s,a) $. The inductive hypothesis is:
    \begin{equation}
        Q^{(n)}(s,a) \leq r(s,a) + \gamma \frac{1-\gamma^n}{1 - \gamma} R.
    \end{equation}
    To prove that the inequality holds for step $(n+1)$, we use the Bellman backup equation:
    \begin{align*}
        Q^{(n+1)}(s,a) &= r(s,a) + \frac{\gamma}{\beta}\E_{s' \sim{} p(\cdot | s,a)} \log \E_{a'\sim{} \pi_0(\cdot|s')} \exp \left( \beta Q^{(n)}(s',a') \right) \\
        Q^{(n+1)}(s,a) &\leq r(s,a) + \frac{\gamma}{\beta}\E_{s' \sim{} p(\cdot | s,a)} \log \E_{a'\sim{} \pi_0(\cdot|s')} \exp \left( \beta \left[ r(s',a') + \gamma \frac{1-\gamma^n}{1 - \gamma}R \right]\right)\\
        &\leq r(s,a) + \gamma \left( R + \gamma \frac{1-\gamma^n}{1 - \gamma}R \right)
    \end{align*}
    At this point if the transition dynamics were known then one could improve this bound by including the next term,
    $\E_{s' \sim{} p(\cdot|s,a)} \max_{a'} r(s',a')$. Instead we do not assume access to the next term, bounding this term by $R$. Then we have:
    % Continuing without this term, we have
    \begin{align*}
        Q^{(n+1)}(s,a) &\leq r(s,a) + \gamma \left( R + \gamma \frac{1-\gamma^n}{1 - \gamma}R \right) \\
        &= r(s,a) + \gamma \frac{1-\gamma^{n+1}}{1 - \gamma}R ,
    \end{align*}
    which completes the proof of the inductive step. As stated above, this completes the proof of the upper bound by taking the limit $n \to \infty$.
    The lower bound follows similarly by swapping all inequalities and taking $\inf$ instead of $\sup$. 
\end{proof}

We now proceed with the proof of our first result, Theorem \ref{thm:gen_double_sided_bounds}. We do so by applying Lemma \ref{lem:double_sided_bound} to the $K^*$ function of \cite{Adamczyk_AAAI}'s Theorem 1. 
\begin{tcolorbox}[colback=blue!5!white,colframe=blue!75!black]
\OnTheFlyBound{appendix}
\end{tcolorbox}
\begin{proof}
    As a point of notation, $\widetilde{r}(s,a)$ in \cite{Adamczyk_AAAI} is the same as our $r(s,a)$. Their $r(s,a)$ is now replaced by the reward function corresponding to an ``optimal'' value function of $Q(s,a)$. As discussed, $Q(s,a)$ need not be an optimal value function corresponding to any known or desired task (reward function). However, because of Theorem~1 in \cite{cao2021identifiability}, we see that choosing a reward function of $Q(s,a)-\gamma \E_{s'} V(s')$ ensures that $Q(s,a)$ is indeed an \textit{optimal} value function, allowing us to apply Theorem 1 of \cite{Adamczyk_AAAI}:
    \begin{equation}
        Q^*(s,a) = Q(s,a) + K^*(s,a)
    \end{equation}
    where $K^*$ is the optimal soft action value function corresponding to a task with reward function $\Delta(s,a) \doteq r(s,a) + \gamma \E_{s' \sim{}p(\cdot|s,a)} V(s') - Q(s,a)$. 
    By applying Lemma~\ref{lem:double_sided_bound} to the value function $K^*$, we arrive at the stated result in Equation~\ref{eq:gen_double_sided_boundappendix}:
    \begin{align*}
        Q^*(s,a) &= Q(s,a) + K^*(s,a) \\
        &\leq Q(s,a) + \Delta(s,a) + \gamma \frac{\sup \Delta}{1-\gamma} \\
        &= Q(s,a) + r(s,a) + \gamma \E_{s' \sim{}p(\cdot|s,a)} V(s') - Q(s,a) + \gamma \frac{\sup \Delta}{1-\gamma} \\
        &= r(s,a) + \gamma \left(\E_{s' \sim{}p(\cdot|s,a)} V(s') + \frac{\sup \Delta}{1-\gamma} \right).
    \end{align*}
    The same proof holds for the lower bound.
\end{proof}

\clearpage
We now turn to the proof of the convergence result presented in the main text:
\begin{tcolorbox}[colback=blue!5!white,colframe=blue!75!black]
\ClippedBOperatorThm{appendix}
\end{tcolorbox}
\begin{proof}
    We first show convergence of the operator ${B}_C$, then show that it converges to the same fixed point as that of $B$. For convergence, it suffices to show that $B_C$ is a contraction mapping, that is: ${|{B}_C Q(s,a) - Q^*(s,a)| \leq~\gamma~|Q(s,a)~-~Q^*(s,a)|}$. % for some $q \in [0,1)$. 
    
    There are three cases for the magnitude of ${B}Q(s,a)$ relative to the upper and lower bounds: 
    \begin{enumerate}
        \item ${B}Q(s,a) \in (L(s,a), U(s,a))$
        \item ${B}Q(s,a) \in (-\infty, L(s,a))$
        \item ${B}Q(s,a) \in (U(s,a), \infty)$
    \end{enumerate}
    
    In the first case, clipping does not occur and hence ${B}_CQ(s,a) = {B}Q(s,a)$, which contracts with rate $\gamma$. %, so $q=\gamma\in [0,1)$.
    In the second case, we can write ${B}Q(s,a) = L(s,a) - \chi(s,a)$ where ${\chi(s,a) := {B}Q(s,a) - L(s,a) > 0}$ is referred to as the ``bound violation''. Then, 
    \begin{align*}
         |{B}_C Q(s,a) - Q^*(s,a)| &= |Q^*(s,a)- {B}_C Q(s,a)|  \\
        &= |Q^*(s,a) - L(s,a)| \\
        &\leq |Q^*(s,a) - L(s,a) + \chi(s,a)|\\
        &= |Q^*(s,a) - (L(s,a) - \chi(s,a))| \\
        &= |Q^*(s,a) - {B}Q(s,a)| \\
        &\leq \gamma |Q(s,a) - Q^*(s,a)|.
    \end{align*} A similar proof holds for the third case.
    
    By the Banach fixed point theorem, it follows that repeated application of ${B}_C$ converges to a fixed point. It is clear that the fixed point for ${B}$ is also a fixed point for ${B}_C$, and since it is unique, we have ${B}_C^\infty Q(s,a) = {B}^\infty Q(s,a) = Q^*(s,a)$.
\end{proof}

\section{Error Analysis for Continuous Spaces\label{app:cont-bounds}}
In this section, we turn to those results specific to the bounds in continuous spaces and their error analysis, based on Lipschitz continuity and finite sampling errors.

\subsection{Reward functions}
In theoretical analyses of RL algorithms it is typical to assume a bounded reward function: $r(s,a)~\in~(R_{\mathrm{min}}, R_{\mathrm{max}})$ for all $s \in \s, a\in \A$. However, the values of these bounds may \textit{not} be known to the agent (or even RL practitioner) in the general model-free case. Thus, one must resort to sampling the reward and estimate the values of $R_{\mathrm{min}}$ and $R_{\mathrm{max}}$. In fact, due to Corollary~\ref{cor:reward-boundsA} from \cite{malherbe2017global} one can obtain global empirical bounds on $r(s,a)$ with high probability. In the following, we restate Corollary~\ref{cor:reward-boundsA} for convenience. Notice that the resulting bounds depend only on a dataset (replay buffer or batch) $\mathcal{B}$, the dimensionality of the continuous state-action space $|\s \times \A|$, a desired probability $1-\delta$, and confidence interval $\varepsilon$. Given these input parameters, the global extrema of the reward function (or any Lipschitz function) can be bounded with high confidence, within the convex hull of points sampled (i.e. the replay buffer). For bounds on the reward function, one obtains the following: %\JA{validity of bounds over convex hull only}

\begin{tcolorbox}[colback=blue!5!white,colframe=blue!75!black]
\RwdCor{A}
\end{tcolorbox}
\begin{proof}
The only difference from the result in \cite{malherbe2017global} is that we have written the result in terms of the number of samples, which is found by solving for $|\mathcal{B}|$:
\begin{align}
    \varepsilon &\leq L_r \cdot \mathrm{diam}(c) \cdot \left(\frac{\log(1/\delta)}{|\mathcal{B}|}\right)^{\frac{1}{d}} \\
    \left|\mathcal{B}\right|&\geq\left( \frac{L_r \mathrm{diam}(c)}{\varepsilon} \right)^{d} \log 1/ \delta.
\end{align}
Note that these bounds only hold within the convex hull of the sampled points. 
\end{proof}

\subsection{Lipschitz Continuity\label{app:lipschitz}}
Due to the hypotheses of Corollary~\ref{cor:reward-boundsA}, the function of interest must be Lipschitz continuous. In the present case, this function (the one being maximized or minimized) is $\Delta$, as seen in Theorem~\ref{thm:gen_double_sided_bounds}. Therefore we must derive the Lipschitz constant for the function $\Delta$. To carry out this calculation, we suppose that a Lipschitz MDP and Lipschitz input function, denoted $Q(s,a)$, are given.
\begin{tcolorbox}[colback=blue!5!white,colframe=blue!75!black]
\begin{customlemma}{B}
Consider an MDP with $(L_r, L_p)$-Lipschitz rewards and dynamics, $L_\kappa$-Lipschitz continuous $\log \pi_0(\cdot | s)$ (with respect to $s$) and $L_Q$-Lipschitz continuous function $Q(s,a)$. The function $\Delta$ in Equation~\ref{eq:exact-delta} generated for this MDP with $Q$ is Lipschitz-continuous with constant
\begin{equation*}
    % L_\Delta = \max\left(L_r, \gamma L_V, L_Q \right),
    L_\Delta = L_r+L_Q + \gamma L_p (L_Q+\beta^{-1}L_\kappa ).
\end{equation*}
In the case of a uniform prior policy $\pi_0$ this simplifies to
\begin{equation}
    L_\Delta=L_r+ (1+ \gamma L_p )L_Q.
\end{equation}
\label{lem:delta-lipschitz}
\end{customlemma}
\end{tcolorbox}
\begin{proof}
The sum of Lipschitz functions is itself Lipschitz continuous, with the Lipschitz constant being the sum of all terms' Lipschitz constants (through the triangle inequality). We begin with the calculation of the Lipschitz constant for the soft state-value function, $V$. First, we note that the operation of LogSumExp is Lipschitz continuous with Lipschitz constant 1 (cf. ``mellowmax'' with an additive constant in \cite{asadi2017alternative}). Since this operation (over action space) is composed with the $L_Q$-Lipschitz input $Q$-function and $L_\kappa$-Lipschitz prior $\log \pi_0$ we have for $V(s')$,
\begin{align}
    \beta^{-1} \log \E_{a'\sim{} \pi_0(\cdot|s')} \exp{\beta Q(s',a')} &=\beta^{-1} \log \sum_a \exp{\beta \left( Q(s',a') + \beta^{-1} \log \pi_0(a'|s')\right)}.
\end{align}
Written this way, we can see the operation in question is a composition and sum of Lipschitz functions, leading to a Lipschitz constant of $L_V = 1\cdot(L_Q+\beta^{-1}L_\kappa)$. Note that in the case of a uniform prior policy $L_\kappa=0$, and the Lipschitz constant for the state value function reduces to $L_Q$. %\JA{For simplicity in the following analysis, we will use this Lipschitz constant.}

Now calculating the Lipschitz constant of $\Delta$, the contribution from $\E_{s' \sim{} p} V(s')$ is:
\begin{align}
    \left| \E_{s'\sim{} p(\cdot|s,a)} V(s') - \E_{{s'}\sim{} p(\cdot|\hat{s},\hat{a})} V({s'}) \right| &= \left| \int_{\s} \left(p(\cdot|s,a) -  p(\cdot|\hat{s},\hat{a}) \right) V({s'}) ds' \right| \\
    &\leq L_p L_V \left( |s-\hat{s}| + |a - \hat{a}|\right).
\end{align}
In the second line we have used the same argument as in the proof of Lemma 2 of \cite{rachelson2010locality}.
Now, using the full definition of $\Delta$ we may finally compute its Lipschitz constant as:
\begin{align*}
    \left| \Delta(s,a) - \Delta(\hat{s}, \hat{a}) \right| &= \left|r(s,a) + \gamma \E_{s'\sim{} p(\cdot|s,a)} V(s') - Q(s,a) - \left( r(\hat{s},\hat{a}) + \gamma \E_{s'\sim{} p(\cdot|\hat{s},\hat{a})} V(s') - Q(\hat{s}, \hat{a}) \right) \right| \\
    &\leq \left(L_r+L_Q\right) \left( \left|s - \hat{s}\right| +  \left|a - \hat{a}\right|\right) + \gamma \left| \E_{s'\sim{} p(\cdot|s,a)} V(s') - \E_{s'\sim{} p(\cdot|\hat{s},\hat{a})} V(s') \right| \\
    &\leq \left(L_r+L_Q\right) \left( \left|s - \hat{s}\right| +  \left|a - \hat{a}\right|\right) + \gamma L_p L_V \left( \left|s - \hat{s}\right| +  \left|a - \hat{a}\right|\right)\\
    &=  \left(L_r+L_Q + \gamma L_p (L_Q+\beta^{-1}L_\kappa )\right) \left( \left|s - \hat{s}\right| +  \left|a - \hat{a}\right|\right).
\end{align*}
The second line follows from the triangle inequality and Lipschitz continuity of the reward function and input function $Q$. This allows us to read off the final Lipschitz constant as:
\begin{equation}
    L_\Delta=L_r+L_Q + \gamma L_p (L_Q+\beta^{-1}L_\kappa ).
\end{equation}
 %in Lemma \ref{lem:entreg_lipschitz_V}. 
Note that in the simpler case of the MaxEnt uniform prior policy ($L_\kappa=0$) the Lipschitz constant of $\Delta$ simplifies to $L_\Delta=L_r+(1+\gamma L_p)L_Q$.
\end{proof}
% \left| \E_{s'\sim{} p(\cdot|s,a)} V(s') - \E_{s'\sim{} p(\cdot|\hat{s},\hat{a})} V(s') \right|
% Now we are positioned to prove Theorem \ref{thm:final-prop}, the double-sided bounds on the soft $Q$-function with sampling errors included.

\subsection{Extension of Exact Bounds}
In this section, we extend our results from the tabular case (Theorem~\ref{thm:gen_double_sided_bounds}) to scenarios where sampling is required (i.e. in the presence of continuous state-action spaces and stochastic dynamics).

We will proceed by introducing and proving three progressively more involved results, covering the following situations: (1) Sampling error arises from estimating the extrema of $\Delta$, which can be calculated exactly, but for which we do not have access to global extrema. (2) An additional sampling error arises due to stochastic transition dynamics. (3) Additional sampling error arises due to continuous action spaces, for which the state-value function integral cannot be calculated exactly.

In each case, we provide the number of samples required for a given $(\varepsilon, \delta)$-concentration inequality. In (1) we denote the number of samples for estimating the extrema of $\Delta$ with $|\mathcal{B}|$ samples (as in practice we sample using the current batch of replay data), and in (2) we introduce $n_\s$, the number of samples for the next-state transitions, and in (3) we introduce $n_\A$, the number of samples for next-actions drawn from the prior policy.

Note that for partially discrete spaces (e.g. continuous state, discrete action), we assume that optimization over the discrete variable is feasible. As in Corollary~\ref{cor:reward-boundsA}, all proceeding bounds involving extremization over $\Delta$ only hold within the convex hull of the sampled points.
% \begin{theorem}
% \JA{need to emphasize that V estimation comes from pi samples}
% (Informal) For sufficiently large $n$ used to estimate $V(s') \approx \hat{V}(s') \doteq \frac{1}{n} \sum_{i=1}^n Q(s,a_i)$ with high probability, 
% \begin{align}
%     Q^*(s,a) &\leq r(s,a)+\gamma \left(\E_{s' \sim{} p} V^\pi(s') + \frac{\sup \hat{\Delta}(s,a)}{1-\gamma} \right) + \mathcal{O} \left( H^2 n^{-1/2}\right)\\
%     Q^*(s,a) &\geq r(s,a)+\gamma \left(\E_{s' \sim{} p} V^\pi(s')  + \frac{\inf \hat{\Delta}(s,a)}{1-\gamma}\right) - \mathcal{O} \left( H^2 n^{-1/2}\right)
% \end{align}
% \end{theorem}

\begin{tcolorbox}[colback=blue!5!white,colframe=blue!75!black]
\begin{theorem}
Consider an MDP with bounded continuous state space, discrete actions, and determinstic dynamics. Let $\varepsilon_1~>~0,\ \delta_1~>~0$, $L_\Delta$ given in Lemma~\ref{lem:delta-lipschitz}, and $\left|\mathcal{B}\right|~\geq~\left( \frac{L_\Delta \mathrm{diam}(c)}{\varepsilon_1} \right)^{|\s|} \log 1/ \delta_1$, be given. Suppose $|\mathcal{B}|$ samples are drawn uniformly from state-space, $s \sim{} \mathrm{Unif}\left(\s\right)$. Then, the following bounds on the $Q$-values
\begin{align}
    Q^*(s,a) &\leq r(s,a)+\gamma \left(\E_{s'\sim{} p} V(s') + \frac{\max_{(s,a) \in \mathcal{B}} {\Delta}(s,a) + \varepsilon_1}{1-\gamma} \right) \\
    Q^*(s,a) &\geq r(s,a)+\gamma \left(\E_{s'\sim{} p} V(s') + \frac{\min_{(s,a) \in \mathcal{B}} {\Delta}(s,a) - \varepsilon_1}{1-\gamma} \right) \ \\
\end{align}
hold with probability at least $1-\delta_1$, where $\Delta$ is given by Equation~\ref{eq:exact-delta}.
\label{thm:deterministic}
\end{theorem}
\end{tcolorbox}

\begin{proof}
The bounds follow directly from applying Corollary~\ref{cor:reward-boundsA} to the bounds given in Theorem~\ref{thm:gen_double_sided_bounds}.
\end{proof}

For the case of stochastic dynamics, we must construct an estimate of $V$ and $\Delta$ which can be calculated with the given information (samples of the next state, rather than an exact integral):
\begin{equation}
    \E_{s'\sim{} p(\cdot|s,a)}{V}(s') =  \int_{\s} p(\cdot|s,a) V(s') ds' \to \frac{1}{n_\s}\sum_{i=1}^{n_\s} V(s_i') \label{eq:vhat-defn}
\end{equation}
\begin{equation}
    \Delta(s,a) \to \hat{\Delta}(s,a) = r(s,a)+\gamma \frac{1}{n_\s}\sum_{i=1}^{n_\s} V(s_i')-Q(s,a)
    \label{eq:deltahat-defn}
\end{equation}
where $n_\s$ denotes the number of next-state samples from the transition dynamics.
Based on these definitions, we introduce two small lemmas, bounding the error in replacing the true functions with their corresponding estimates:

\begin{tcolorbox}[colback=blue!5!white,colframe=blue!75!black]
\begin{customlemma}{C}
Let $\delta>0$ and $\varepsilon>0$ be given. Then with at least $n\geq \left( \frac{R_\mathrm{max}-R_\mathrm{min}}{\varepsilon(1-\gamma)}\right)^2 \log \frac{2}{\delta}$ samples on the next state, the error in replacing $\E_{s'} V(s')$ with $\frac{1}{n}\sum_{i=1}^n V(s_i')$ is bounded with probability $1-\delta$, leading to the following bounds:
\begin{align}
    \E_{s'\sim{}p} V(s') &\leq \frac{1}{n} \sum_{i=1}^{n} V(s_i') +  \varepsilon \\
    \E_{s'\sim{}p} V(s') &\geq \frac{1}{n} \sum_{i=1}^{n} V(s_i') - \varepsilon.
\end{align}
% which hold with probability $1-\delta$.
\label{lem:next-state}
\end{customlemma}
\end{tcolorbox}
\begin{proof}
Applying Hoeffding's inequality on the relevant term gives
\begin{align*}
    \mathbb{P} \left(\left|\E_{s'\sim{}p} V(s') - \frac{1}{n} \sum_{i=1}^{n} V(s_i')\right| < \varepsilon \right) &\geq 1 - 2 \exp\left(-\frac{2\varepsilon^2}{b^2}n\right)
\end{align*}
where as usual $b=\left(R_{\text{max}} - R_{\text{min}}\right) (1-\gamma)^{-1}$ is the gap between a lower and upper bound on the concentrating quantity of interest, $V$. Note that here and in the following we assume exact global bounds on the reward function, $r(s,a)$ though in principal a corresponding error term based on finite samples can be included. Defining $\delta=2\exp\left(-\frac{2\varepsilon^2}{b^2}n\right) $, we have with probability at least $1-\delta$:
\begin{align}
    \left|\E_{s'\sim{}p} V(s') - \frac{1}{n} \sum_{i=1}^{n} V(s_i')\right| &< \varepsilon, % 2\frac{R_{\text{max}} - R_{\text{min}}}{1-\gamma} \sqrt{\frac{1}{2n}\log \frac{2}{\delta_2} } \doteq \varepsilon \\
    % \E_{s'\sim{}p} V(s') &\leq \frac{1}{n} \sum_{i=1}^{n} V(s_i') +  \varepsilon_2,
\end{align}

where solving for $n$ yields a requirement of $n=\frac{1}{2} \left( \frac{R_\mathrm{max}-R_\mathrm{min}}{\varepsilon(1-\gamma)}\right)^2 \log \frac{2}{\delta}$ samples. Expanding the absolute values leads to the two bounds shown, which is a more useful form for the subsequent results.
\end{proof}

We next provide a similar lemma for the error in replacing $\Delta$ with $\hat{\Delta}$:
\begin{tcolorbox}[colback=blue!5!white,colframe=blue!75!black]
\begin{customlemma}{D}
Let $\varepsilon>0$ and $\delta>0$ be given. Then given $n \geq \frac{1}{2}\left( \frac{R_\mathrm{max}-R_\mathrm{min}}{\varepsilon(1-\gamma)}\right)^2 \log \frac{2}{\delta}$ samples to estimate the value of the next state, the error in replacing $\Delta(s,a)$ with $\hat{\Delta}(s,a)$ in Equation~\ref{eq:deltahat-defn} is upper bounded. That is, for all $s\in\s, a\in\A$:
\begin{equation}
    \left|\Delta(s,a) - \hat{\Delta}(s,a)\right| \leq \gamma \varepsilon %\left|\E_{s'\sim{} p(\cdot|s,a)} V(s') - \frac{1}{n} \sum_{i=1}^n V(s_i')\right|
\end{equation}
with probability $1-\delta$.
\end{customlemma}
\end{tcolorbox}
\begin{proof}
From the definitions, we can immediately calculate the following error bound
\begin{align*}
    \left|\Delta(s,a) - \hat{\Delta}(s,a)\right| &= \left|r(s,a) + \gamma \E_{s'\sim{}p} V(s') - Q(s,a) - \left( r(s,a) + \gamma \frac{1}{n} \sum_{i=1}^n V(s_i') - Q(s,a) \right) \right| \\
    &=\gamma \left|\E_{s'\sim{}p} V(s') - \frac{1}{n} \sum_{i=1}^n V(s_i')\right| \\ 
    &\leq \gamma \varepsilon.
\end{align*}
where the last line holds with probability $1-\delta$. The last line follows from Lemma~\ref{lem:next-state} when using at least $n= \frac{1}{2}\left( \frac{R_\mathrm{max}-R_\mathrm{min}}{\varepsilon(1-\gamma)}\right)^2 \log \frac{2}{\delta}$ samples.
\end{proof}

Now we combine all the previous results to arrive at the following extension of our main results to the case of sampling in stochastic environments with continuous state space:
\begin{tcolorbox}[colback=blue!5!white,colframe=blue!75!black]
\begin{theorem}
Consider an MDP with bounded continuous state space, discrete actions and stochastic dynamics.  Let $\varepsilon_1, \varepsilon_2~>~0,\ \delta_1, \delta_2~>~0$, and $\left|\mathcal{B}\right|~\geq~\left( \frac{L_\Delta \mathrm{diam}(c)}{\varepsilon_1} \right)^{|\s|} \log 1/\delta_1$, $n_\s~\geq~\frac{1}{2} \left( \frac{R_\mathrm{max}-R_\mathrm{min}}{\varepsilon_2(1-\gamma)}\right)^2 \log \frac{2}{\delta_2}$, be given. Suppose $|\mathcal{B}|$ samples are drawn uniformly from state-space, $s \sim{} \mathrm{Unif}\left(\s\right)$. Then, for ${V},\ \hat{\Delta}$ given in Equation~\ref{eq:vhat-defn}, \ref{eq:deltahat-defn}, the following bounds on the $Q$-values
\begin{align}
    Q^*(s,a) &\leq r(s,a)+\gamma \left(\frac{1}{n_\s}\sum_{i=1}^{n_\s}{V}(s'_i) + \frac{\max_{(s,a) \in \mathcal{B}} \hat{\Delta}(s,a)  + 
    \varepsilon_1 + \varepsilon_2}{1-\gamma} \right) \\
    Q^*(s,a) &\geq r(s,a)+\gamma \left( \frac{1}{n_\s}\sum_{i=1}^{n_\s}{V}(s'_i) + \frac{\min_{(s,a) \in \mathcal{B}} \hat{\Delta}(s,a) -
    \varepsilon_1 - \varepsilon_2}{1-\gamma} \right)
\end{align}
hold with probability at least $1-\delta_1 - 2\delta_2$. 
\label{thm:stochastic}
\end{theorem}
\end{tcolorbox}

\begin{proof}
For stochastic dynamics there remains an error in using samples to estimate the expectation over next states in $V(s')$.
% To this end, let us use Hoeffding's inequality on the relevant term, which states
% \begin{align}
%     \mathbb{P} \left(\left|\E_{s'\sim{}p} V(s') - \frac{1}{n} \sum_{i=1}^{n} V(s_i')\right| < \varepsilon_2 \right) &\geq 1 - 2 \exp\left(-\frac{2\varepsilon_2^2}{b^2}n\right) \\
% \end{align}
% where as usual $b=\left(R_{\text{max}} - R_{\text{min}}\right) (1-\gamma)^{-1}$ is the gap between a lower and upper bound on the concentrating quantity of interest, $V$. Equivalently, with probability at least $1-\delta_2$, we have
% \begin{align}
%     \left|\E_{s'\sim{}p} V(s') - \frac{1}{n} \sum_{i=1}^{n} V(s_i')\right| &< 2\frac{R_{\text{max}} - R_{\text{min}}}{1-\gamma} \sqrt{\frac{1}{2n}\log \frac{2}{\delta_2} } \doteq \varepsilon_2 \\
%     \E_{s'\sim{}p} V(s') &\leq \frac{1}{n} \sum_{i=1}^{n} V(s_i') +  \varepsilon_2,
% \end{align}

% where solving for $n_\s$ yields a requirement of $n_\s=\frac{HR_{\text{max}} - R_{\text{min}}}{4\varepsilon_2^2}$ samples.

Now, for simplicity, we will assume that the same number of samples $n_\s$ are used to estimate $V(s')$ appearing explicitly in the bound \textit{and} implicitly in the definition of $\Delta$. In principle, these can be different values, but we propose (as done experimentally) to use the same batch of replay data for both calculations.

% To propagate such an error in $\Delta$, we define a new function $\hat{\Delta}$ in Equation~\ref{eq:deltahat-defn}. Notice that this will also affect the value of the estimated maximum and minimum of $\hat{\Delta}$ in the following way:
% \begin{align}
%     \left| \Delta(s,a) - \hat{\Delta}(s,a)\right| &= \gamma  \left|\E_{s'\sim{}p} V(s') - \frac{1}{n} \sum_{i=1}^{n} V(s_i')\right| \\
%     &\leq \gamma \varepsilon_2
% \end{align}
% with probability $1-\delta_2$.
Then by Lemma C, let $\varepsilon_2>0, \delta_2>0$ and $n_\s \geq  \frac{1}{2}\left( \frac{R_\mathrm{max}-R_\mathrm{min}}{\varepsilon_2(1-\gamma)}\right)^2 \log \frac{2}{\delta_2}$ next-state samples be given. Then with probability at least $1-\delta_2$:
\begin{align}
    \Delta(s,a) &\leq \hat{\Delta}(s,a) + \gamma \varepsilon_2 \\
    \max_{(s,a) \in \mathcal{B}} \Delta(s,a) &\leq \max_{(s,a) \in \mathcal{B}} \left(\hat{\Delta}(s,a) + \gamma \varepsilon_2  \right)\\
    &\leq \max_{(s,a) \in \mathcal{B}} \hat{\Delta}(s,a) + \gamma \varepsilon_2.
\end{align}

Combining the bound above with Lemma B allows one to replace all instances of $V$ and $\Delta$ with their approximations in the upper bound of Theorem~\ref{thm:deterministic}:
\begin{align*}
     Q^*(s,a) &\leq r(s,a)+\gamma \left(\E_{s'\sim{} p} V(s') + \frac{\max_{(s,a) \in \mathcal{B}} {\Delta}(s,a) + \varepsilon_1}{1-\gamma} \right) \\
     &\leq r(s,a)+\gamma \left(\frac{1}{n_\s}\sum_{i=1}^{n_\s}{V}(s') +  \varepsilon_2 + \frac{\max_{(s,a) \in \mathcal{B}} {\Delta}(s,a) + \varepsilon_1}{1-\gamma} \right) \\
     &\leq r(s,a)+\gamma \left(\frac{1}{n_\s}\sum_{i=1}^{n_\s}{V}(s') +  \varepsilon_2 + \frac{\max_{(s,a) \in \mathcal{B}} \hat{\Delta}(s,a) + \gamma \varepsilon_2 + \varepsilon_1}{1-\gamma} \right) \\
     &= r(s,a)+\gamma \left(\frac{1}{n_\s}\sum_{i=1}^{n_\s}{V}(s') +  \frac{\max_{(s,a) \in \mathcal{B}} \hat{\Delta}(s,a) + \varepsilon_1 + \varepsilon_2}{1-\gamma} \right),
\end{align*}
which holds with probability at least $(1-\delta_1)(1-\delta_2)^2\geq 1-\delta_1-2\delta_2$ (we ignore the terms beyond first order which are negligible in the limit of small $\delta_i$). One factor of $1-\delta_1$ arises from Theorem~\ref{thm:deterministic} and two factors of $1-\delta_2$ correspond to the use of Lemma B and Lemma C separately, as they act on independent next-state samples: Lemma B operates on the next-state samples dictated by the fixed $(s,a)$-value of interest on the left-hand side of the bound, whereas Lemma C operates on the next states in the batch $\mathcal{B}$. As discussed previously, we have assumed for simplicity the same values of $(\varepsilon_2, \delta_2)$ in their corresponding concentration bounds.

A similar proof holds for the lower bound.
\end{proof}

Lastly, we will consider the case of continuous actions in a soft Q-learning style algorithm where one samples actions from the prior policy $\pi_0$ to estimate the following integral
\begin{equation}
    V(s) = \beta^{-1} \log \int_{\A} e^{\beta Q(s,a)} \pi_0(a|s) da,
    \label{eq:V-exact}
\end{equation}
specifically with $n_\A$ samples from the prior policy,
\begin{equation}
    \hat{V}(s) \doteq \beta^{-1} \log \sum_{i=1}^{n_\A} e^{\beta Q(s,a_i)}.
    \label{eq:V-approx}
\end{equation}

\begin{tcolorbox}[colback=blue!5!white,colframe=blue!75!black]
\FullPAC{A}
\end{tcolorbox}

As before, we first provide a bound on the error in replacing $V$ with $\hat{V}$ before combining it with the previous result. 
% \JA{tighten this with the policy variance in Gaussian case}
\begin{tcolorbox}[colback=blue!5!white,colframe=blue!75!black]
\begin{customlemma}{E}
% Let the state value and estimated state value be defined as

% \begin{align}
%     V(s) &= \frac{1}{\beta} \log \E_{a \sim{} \pi_0(a|s)} e^{\beta Q(s,a} \\
%     \hat{V}(s) &= \frac{1}{\beta} \log \sum_{i=1}^{n_a} e^{\beta Q(s,a_i)}
% \end{align}
Let the definitions in Equations~\ref{eq:V-exact}, \ref{eq:V-approx} and some $\varepsilon >0,\ \delta >0$ be given.
Then, for
\begin{equation}
    n \geq  \frac{1}{2} \left( \frac{e^{\beta (R_{\mathrm{max}}-R_{\mathrm{min}})/(1-\gamma)}-1}{e^{\beta \varepsilon} -1 } \right)^2 \log\frac{2}{\delta} 
\end{equation}
action samples from the prior policy $a \sim{} \pi_0(a | s)$, the error in replacing the state value function with its estimate is bounded:
\begin{equation}
     \frac{1}{n} \sum_{i=1}^{n} e^{\beta Q(s,a_i)} -\varepsilon < V(s) < \frac{1}{n} \sum_{i=1}^{n} e^{\beta Q(s,a_i)} + \varepsilon
\end{equation}
with probability $1-\delta$.
\label{lem:action-sampling}
\end{customlemma}
\end{tcolorbox}

\begin{proof}
We will focus on the proof of the upper bound. The lower bound follows from a similar proof. From Hoeffding's inequality, with probability $1-\delta$,
\begin{align*}
    \left| e^{\beta V(s)} - \frac{1}{n} \sum_{i=1}^{n} e^{\beta Q(s,a_i)} \right| &< \left( e^{\beta R_\mathrm{\max} / (1-\gamma)} - e^{\beta R_\mathrm{\min} / (1-\gamma)} \right) \sqrt{\frac{1}{2n} \log\frac{2}{\delta}} \doteq \widetilde{\varepsilon} \\
    e^{\beta V(s)} &\leq e^{\beta \hat{V}(s)} + \widetilde{\varepsilon}.
\end{align*}
By taking the $\log$ on both sides
\begin{align*}
    V(s) &\leq \frac{1}{\beta} \log( e^{\beta \hat{V}(s)} + \widetilde{\varepsilon}) \\
    &= \hat{V}(s) + \beta^{-1} \log( 1 + \widetilde{\varepsilon} e^{-\beta \hat{V}(s)}) \\
    &\leq \hat{V}(s) + \beta^{-1}\log\left( 1+  \widetilde{\varepsilon} e^{-\beta R_{\mathrm{min}}/(1-\gamma) } \right)  \\
    &\doteq \hat{V}(s) + \varepsilon_+
\end{align*}
(with probability $1-\delta$),
where $\varepsilon_+$ satisfies 
\begin{equation}
    \frac{e^{\beta \varepsilon_+} - 1}{e^{-\beta R_\mathrm{min} / (1-\gamma)}} = \widetilde{\varepsilon} = \left( e^{\beta R_\mathrm{\max} / (1-\gamma)} - e^{\beta R_\mathrm{\min} / (1-\gamma)} \right) \sqrt{\frac{1}{2n_+} \log\frac{2}{\delta}}.
    \label{eq:eps+}
\end{equation}
or in other words,
\begin{equation}
    n_+ = \frac{1}{2} \log\frac{2}{\delta} \left( \frac{e^{\beta (R_{\mathrm{max}}-R_{\mathrm{min}})/(1-\gamma)}-1}{e^{\beta \varepsilon_+} - 1} \right)^2.
\end{equation}

For clarity we also provide the corresponding lower bound here:
\begin{align*}
    e^{\beta V(s)} &\geq e^{\beta \hat{V}(s)}-\widetilde{\varepsilon} \\
    V(s) &\geq \frac{1}{\beta} \log\left( e^{\beta \hat{V}(s)} -\widetilde{\varepsilon}  \right) \\
    &= \hat{V}(s) + \beta^{-1} \log( 1 - \widetilde{\varepsilon} e^{-\beta \hat{V}(s)}) \\
    & \geq \hat{V}(s) + \beta^{-1} \log( 1 - \widetilde{\varepsilon} e^{-\beta R_{\mathrm{max}}/(1-\gamma)}) \\
    &\doteq \hat{V}(s) - \varepsilon_-
\end{align*}
where again we solve for $n_-$:
\begin{equation}
    n_- = \frac{1}{2} \log\frac{2}{\delta} \left( \frac{1-e^{-\beta (R_{\mathrm{max}}-R_{\mathrm{min}})/(1-\gamma)}}{1-e^{-\beta \varepsilon_-} } \right)^2.
\end{equation}

Letting $\varepsilon_+ = \varepsilon_- \doteq \varepsilon$ for simplicity and solving for the value of $n$ in Hoeffding's inequality such that both the upper and lower bounds on $V$ are satisfied gives:
\begin{align}
    n &\geq \max \left\{n_+, n_- \right\} \\
    &= \frac{1}{2} \log\frac{2}{\delta}  \max \left\{  \left( \frac{e^{\beta (R_{\mathrm{max}}-R_{\mathrm{min}})/(1-\gamma)}-1}{e^{\beta \varepsilon} -1 } \right)^2, \left( \frac{1-e^{-\beta (R_{\mathrm{max}}-R_{\mathrm{min}})/(1-\gamma)}}{1-e^{-\beta \varepsilon} } \right)^2  \right\}.
\end{align}
For small $\varepsilon < (R_{\mathrm{max}}-R_{\mathrm{min}})/(1-\gamma)$, the first term dominates, thus
\begin{equation}
    n \geq  \frac{1}{2} \log\frac{2}{\delta}  \left( \frac{e^{\beta (R_{\mathrm{max}}-R_{\mathrm{min}})/(1-\gamma)}-1}{e^{\beta \varepsilon} -1 } \right)^2
\end{equation}
samples suffice to satisfy both the lower and upper bounds $|V-\hat{V}| < \varepsilon$ with probability at least $1-\delta$. In passing, we note that in the low $\beta$ regime, $\beta \varepsilon \ll 1$ and $\beta (R_{\mathrm{max}}-R_{\mathrm{min}})/(1-\gamma)\ll 1$, the required samples simplifies to the ``usual'' (Hoeffding) form, quadratic in $H/\varepsilon$:
\begin{equation}
     n(\beta \ll 1) \geq \frac{1}{2}\left( \frac{H(R_\mathrm{max}-R_\mathrm{min})}{\varepsilon}\right)^2 \log \frac{2}{\delta}.
\end{equation}

% \JA{Question: Can we combine all of these to say, with probability $1-\delta_1-\delta_2-\delta_3$,

% \begin{align}
%     Q^*(s,a) &\leq r(s,a)+\gamma \left(\frac{1}{n} \sum_{i=1}^{n} \hat{V}(s_i') + \frac{\max_{(s,a) \in \mathcal{B}} \hat{\Delta}(s,a)}{1-\gamma} \right) + \gamma \frac{\varepsilon_1 + \varepsilon_2 + \varepsilon_3 e^{-\beta R_{\mathrm{min}/(1-\gamma)}}}{1-\gamma} \\
% \end{align}
% where the hat on $\Delta$ now represents the calculation with both uncertain values, thus giving the $\varepsilon_3$ dependence outside.
% }
\end{proof}

Now, to prove Theorem~\ref{thm:final-prop-formalA}, we apply the same techniques as before:

\begin{proof}
Applying Lemma~\ref{lem:action-sampling} to the expectation over actions in $\hat{V}$ and $\hat{\Delta}$ in Theorem~\ref{thm:stochastic} gives, similar to the previous proof, two terms of $\varepsilon_3$:

\begin{align}
    Q^*(s,a) &\leq r(s,a)+\gamma \left(\frac{1}{n_\s}\sum_{i=1}^{n_\s}{V}(s') + \frac{\max_{(s,a) \in \mathcal{B}} \hat{\Delta}(s,a)  + 
    \varepsilon_1 + \varepsilon_2}{1-\gamma} \right) \\
    &\leq r(s,a)+\gamma \biggr(\frac{1}{n_\s}\sum_{i=1}^{n_\s}\hat{V}(s') + \varepsilon_3 +
    \frac{\max_{(s,a) \in \mathcal{B}} \hat{\Delta}(s,a) + \varepsilon_1 + \varepsilon_2+\gamma \varepsilon_3 }{1-\gamma} \biggr) \\
    &= r(s,a)+\gamma \biggr(\frac{1}{n_\s}\sum_{i=1}^{n_\s}\hat{V}(s') + \frac{\max_{(s,a) \in \mathcal{B}} \hat{\Delta}(s,a) + \varepsilon_1 + \varepsilon_2+\varepsilon_3}{1-\gamma} \biggr).
\end{align}
Similar to the proof of Theorem~\ref{thm:stochastic}, we introduced two instances of this action sampling (one for $V(s')$ and one for the extrema of $\hat{\Delta}(s,a)$). This requires an additional two factors of $1-\delta_3$ in the confidence:
$(1-\delta_1)(1-\delta_2)^2(1-\delta_3)^2 \geq 1-\delta_1-2\delta_2-2\delta_3.$
\end{proof}

We note that one can also instead combine Theorem~\ref{thm:deterministic} with Lemma~\ref{lem:action-sampling} to arrive at double-sided bounds for the case of deterministic transitions with continuous actions.
\end{document}